\documentclass[lettersize,journal]{IEEEtran}
\usepackage{amsmath,amsfonts}
\usepackage{algorithmic}
\usepackage{array}
\usepackage[caption=false,font=scriptsize,labelfont=rm,textfont=rm]{subfig}

\usepackage{textcomp}
\usepackage{stfloats}
\usepackage{url}
\usepackage{verbatim}
\usepackage{graphicx}

\usepackage{multirow}
\usepackage{booktabs}
\usepackage{enumitem}
\usepackage{color}
\usepackage{subfig}
\usepackage{float}
\usepackage{wrapfig}
\usepackage{tabularx}
\usepackage{mathbbol}
\usepackage{amsmath}
\usepackage{amssymb}
\usepackage[table]{xcolor}
\usepackage{colortbl}
\graphicspath{{figures/}}

\usepackage{tikz}
\definecolor{my_color_yellow}{RGB}{255, 255, 200} 
\definecolor{my_color_blue}{RGB}{200, 230, 255} 
\definecolor{my_color_green}{RGB}{200, 255, 200} 
\usepackage{soulutf8}

\sethlcolor{white} 

\def\BibTeX{{\rm B\kern-.05em{\sc i\kern-.025em b}\kern-.08em
    T\kern-.1667em\lower.7ex\hbox{E}\kern-.125emX}}
\usepackage{balance}
\begin{document}
\title{DiCLIP: Diffusion Model Enhances CLIP's Dense Knowledge for Weakly Supervised Semantic Segmentation}
\author{Zhiwei Yang, Pengfei Song, Yucong Meng, Kexue Fu, Shuo Wang, Zhijian Song
\thanks{Zhiwei Yang is with Academy for Engineering and Technology, Fudan University, Shanghai 200433, China. Zhijian Song is with Zhongshan Hospital, Fudan University, Shanghai 20032, P.R. China. Yucong Meng, Shuo Wang, and Zhijian Song are with Digital Medical Research Center, School of Basic Medical Science, Fudan University, and also with Shanghai Key Lab of Medical Image Computing and Computer Assisted Intervention, Shanghai 200032, China. Kexue Fu and Pengfei, Song are with Shandong Computer Science Center, China. Shuo Wang and Zhijian Song are the corresponding authors. (email: shuowang@fudan.edu.cn, zjsong@fudan.edu.cn).

This work is supported by National Natural Science Foundation of China (62501340), Taishan Scholars Program (NO.tsqn202408245), the Linyi People's Hospital under Grant No.2024LYKC002, Shanghai QiYuan Innovation Foundation, and International Science and Technology Cooperation Program under the 2023 Shanghai Action Plan for Science (23410710400).}}

\markboth{Journal of \LaTeX\ Class Files,~Vol.~18, No.~9, March~2025}%
{How to Use the IEEEtran \LaTeX \ Templates}

\maketitle

\begin{abstract}
   Weakly Supervised Semantic Segmentation (WSSS) with image-level labels typically leverages Class Activation Maps (CAMs) to achieve pixel-level predictions. Recently, Contrastive Language-Image Pre-training (CLIP) has been introduced to generate CAMs in WSSS. However, previous WSSS methods solely adopt CLIP's vision-language paired property for dense localization, neglecting its inherently limited dense knowledge across both visual and text modalities, which renders CAM generation suboptimal. In this work, we propose DiCLIP, a novel WSSS framework that leverages the generative diffusion model to enhance CLIP's dense knowledge across two modalities. Specifically, Visual Correlation Enhancement (VCE) and Text Semantic Augmentation (TSA) modules are proposed for dense prediction enhancement. To improve the spatial awareness of visual features, our VCE module utilizes diffusion's reliable spatial consistency to mitigate the over-smoothing issue in CLIP's attention. It designs the Attention Clustering Refinement (ACR) module to reliably extract diverse correlation maps from the diffusion model. The correlation maps act as a diversity bias for CLIP's self-attention, recursively pushing its visual features towards a more discriminative dense distribution. To augment the semantics of text embeddings, our TSA module argues that a single text modality is insufficient to encompass the variability of visual categories. Thus, we leverage diffusion's generative power to maintain a dynamic key-value cache model, shifting CAM generation from a patch-text matching mechanism to a novel visual knowledge retrieval paradigm. With these enhancements, DiCLIP not only outperforms state-of-the-art methods on PASCAL VOC and MS COCO but also significantly reduces training costs. Code is publicly available at https://github.com/zwyang6/DiCLIP.
\end{abstract}


\section{Introduction}

\IEEEPARstart{S}{emantic} segmentation intends to classify every pixel of an input image into different semantic groups, which has achieved significant success across computer vision scenarios~\cite{f1,segformer,mambaseg,tip1_fss,tip2_fss}. Previous literature heavily relies on pixel-level annotations on large-scale datasets. This fully supervised setting involves intensive human labor, hindering its extensive applications. To alleviate this shortcoming, weakly-supervised semantic segmentation (WSSS) has appeared as an efficient alternative for dense predictions. It enables pixel-wise predictions with more accessible annotations like points~\cite{1}, scribbles~\cite{2,3}, bounding boxes~\cite{4,5}, or image-level labels~\cite{more,7,12}. Among all these low-cost annotation types, most WSSS methods utilize image-level labels to provide dense localization cues, connecting visual concepts with pixel-level semantics~\cite{6,8}. In this work, we investigate WSSS with image-level labels as well.
\begin{figure}[t]
  \centering
  \includegraphics[width=8.7cm]{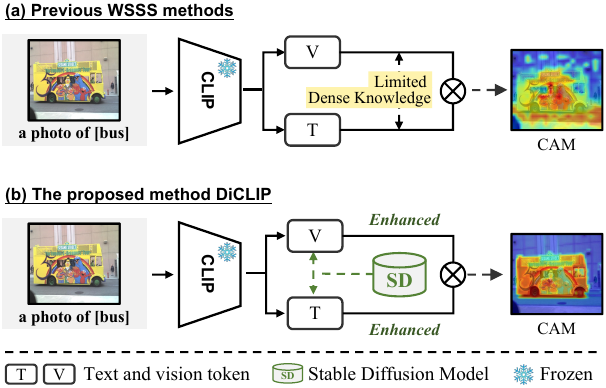}
   \caption{Motivation of DiCLIP. (a) Previous WSSS methods solely rely on CLIP's vision-text paired property for dense localization, while ignoring its inherently limited dense knowledge across both visual and text modalities, resulting in suboptimal CAM generation. (b) The proposed DiCLIP harnesses the powerful Stable Diffusion (SD)~\cite{sd} to enhance CLIP's dense knowledge across two modalities, generating more precise and complete CAMs.}
   \label{fig.1}
   \vspace{-1.5em}
\end{figure}

Most WSSS methods with image-level annotations consist of three phases: $1)$ initially generating Class Activation Maps (CAMs)~\cite{11} from a well-trained classification model; $2)$ refining CAMs into pseudo labels~\cite{12,t1,t3}; $3)$ training a segmentation model with dense supervision from pseudo labels~\cite{14}. Commonly, the main challenge of WSSS lies in how to generate high-quality CAMs. However, due to the minimal semantic information in image-level labels, CAMs tend to highlight the most distinctive object parts, significantly hindering WSSS performance~\cite{9}. Recently, Contrastive Language-Image Pre-training (CLIP)~\cite{14}, known for pre-training on $400$ million image-text pairs, has been applied to generate CAMs in WSSS. In general, there are two CLIP-based paradigms for CAM generation, i.e., image-text alignment and patch-text alignment. The former uses well-aligned global image-text embeddings to regularize different visual concepts~\cite{15} or generates high-quality GradCAMs by computing the gradient from cross-modality matching~\cite{17,18}. In contrast, the latter~\cite{weakclip} produces visual responses by calculating the patch-wise similarities between text embeddings and every patch token, yielding better activations.
\hl{Although the patch–text alignment appears more suitable for dense prediction than image-text alignment, CLIP's image-level training paradigm fundamentally limits the semantic granularity of its visual and textual representations. As a result, such alignment remains inherently constrained. Existing WSSS methods rely solely on CLIP's inherent vision-language alignment and therefore fail to overcome this limitation, leading to suboptimal CAM generation, as shown in Fig.~{\ref{fig.1}} (a).}

In this work, we propose DiCLIP, a novel patch-text aligned framework that leverages the powerful generative diffusion model (Stable Diffusion~\cite{sd}, SD) to enhance CLIP's dense knowledge, as shown in Fig.~\ref{fig.1} (b). Our findings reveal that the off-the-shelf spatial consistency and generative power of diffusion models effectively boost CLIP's CAM generation. To achieve this, we introduce Visual Correlation Enhancement (VCE) and Text Semantic Augmentation (TSA) modules, which transfer dense knowledge from the diffusion model to CLIP's vision and text modalities, respectively.

To improve the spatial awareness of visual features, we propose the VCE module to utilize diffusion's reliable spatial consistency in a non-parametric manner. We observe that CLIP's over-smoothed attention degrades fine-grained features, whereas diffusion models preserve diverse semantic representations (illustrated in Fig.~\ref{fig.3}). Motivated by it, we extract attention maps from diffusion models, leveraging them to enhance the diversity of CLIP's visual features. To further improve the quality of extracted attention maps, we design the Attention Clustering Refinement (ACR) module for better visual compensation. ACR clusters the attention maps into different semantic-relevant regions and captures token affinities among grouped semantics. These affinities are used to recursively refine diffusion's attention maps, improving correlations within the same semantics while suppressing irrelevant regions. The refined attentions are then used to calibrate CLIP's attention from intermediate layers, progressively diversifying its over-smoothed features without any training process.

To enrich the semantics of text embeddings, our TSA module argues that a single text modality is insufficient to encompass the variability of visual categories. Thus, we leverage diffusion's generative power to maintain a key-value cache model~\cite{unboundcachemodel,tipadapter}, shifting CAM generation from patch-text alignment to a novel visual knowledge retrieval paradigm. Specifically, we generate single-class images from the Stable Diffusion model and encode them into positive/negative feature embeddings via CLIP's visual encoder, forming Keys in the cache model. We utilize text embeddings to weight one-hot class labels, storing them as Values to capture intra- and inter-class differences among cached Keys. With the cache model, TSA first enhances text modality through static knowledge retrieval, which generates high-quality CAMs in a training-free manner. In addition, a learnable adapter is further built with the key-value pairs, enabling optimized CAM generation via dynamic knowledge retrieval. These retrieval processes effectively utilize diffusion's visual knowledge to enrich CLIP's text representation for CAM generation.

In summary, our main contributions are listed as follows:
\begin{itemize}
    \item We propose a novel patch-text aligned WSSS framework DiCLIP, that leverages diffusion model (Stable Diffusion) to enhance CLIP's dense ability across visual and text modalities. DiCLIP efficiently generates high-fidelity CAMs in both training-free and training-efficient manners, demonstrating its strong potential in WSSS.
    \item To improve the spatial awareness of visual features, the proposed Visual Correlation Enhancement (VCE) module utilizes diffusion's reliable spatial consistency to calibrate CLIP's smoothing attention, effectively diversifying the visual features without any training. 
    \item To enrich the semantics of text embeddings, the designed Text Semantic Augmentation (TSA) module leverages diffusion's generative power to maintain a visual key-value cache model, pioneering a novel patch-wise knowledge retrieval paradigm for CAM generation.
    \item Extensive experiments are conducted on PASCAL VOC 2012 and MS COCO 2014, strongly demonstrating that DiCLIP holds significant superiority over previous state-of-the-art methods with less training cost.
\end{itemize}

\section{Related Work}
\subsection{Weakly-supervised Semantic Segmentation}
{In general, the pipeline of WSSS is divided into generating CAMs from well-trained classification models, refining CAMs into pseudo labels, and retraining a segmentation network~\cite{tip3,tip4,PPC,32}. With this pipeline, WSSS can be further classified into multi-stage and single-stage paradigms. In multi-stage methods~{\cite{t4,t5, PLDA, tip2}}, the classification and segmentation networks are trained separately. As a result, they intend to hold superior performance with a complicated and time-consuming training process.
The single-stage approaches~{\cite{more,seco,12,22}} train the classification and segmentation models in parallel. Both models share the same encoder, enabling an end-to-end pipeline for CAM generation and segmentation. These methods commonly yield inferior performance, but are more efficient to implement. In this paper, we define DiCLIP as a single-stage framework, as it generates CAMs and segmentation maps in parallel during training. {To avoid repeatedly generating images for each training, we precompute the key-value cache model.} For fairness, we compare DiCLIP with both single-stage and multi-stage methods.}

For both single-stage and multi-stage paradigms, generating high-quality CAMs is the most fundamental step. {GSM{~\cite{GSM}} proposes a graph neural network for group-wise semantic mining and views input images as graph nodes for strong performance. PGSeg{~\cite{PGseg}} enhances plain ViT by incorporating grouping recognition. It introduces a non-learnable prototypical regularization for both image and text modalities, thereby achieving better predictions. In contrast to this type, another line of research leverages the vision–language model CLIP to generate CAMs.} CLIMs~\cite{15} uses image-text pairs of CLIP to regularize different concepts of CAMs. CLIP-ES~\cite{16} and WeCLIP~\cite{18} use image-text matching to generate GradCAM. Apart from these efforts, a few attempts are made to improve the image-text alignment by enriching CLIP's text templates~\cite{qaclims,promptclass}. In contrast to the image-text paradigm, WeakCLIP~\cite{weakclip} investigates the patch-text alignment to generate CAMs, which demonstrates better performance. 

\hl{In this work, we also adopt patch–text alignment, but differ fundamentally from existing CLIP-based WSSS methods in both modeling assumptions and inference strategy. Prior approaches rely solely on CLIP’s inherent vision–language alignment (e.g., WeCLIP~{\cite{18}}, WeakCLIP~{\cite{weakclip}}), implicitly assuming that CLIP provides sufficiently dense semantic representations for patch-level supervision. In contrast, we explicitly identify dense semantic deficiency as a fundamental limitation of CLIP, which leads to unreliable patch–text alignment and CAM quality. To address this, we introduce diffusion models as an external source of visual knowledge. Conceptually, our method reformulates the patch–text alignment into a visual knowledge retrieval process, enabling enhancement of CLIP’s dense representations and more accurate CAM generation.}
\vspace{-1em}
\subsection{Text-to-image Diffusion Model}
Text-to-image diffusion models have seen tremendous success due to their powerful generative ability conditioned on texts. Among these, Stable Diffusion (SD)~\cite{sd} is one of the most prominent text-to-image generators. It works efficiently based on a VAE latent space, with a denoised Transformer-based UNet~\cite{UNet} incorporated for text conditioning. Recently, inspired by its reliable spatial representation, many methods have applied it for segmentation tasks. VPD~\cite{VPD} uses SD as a feature extractor and fine-tunes it for fully-supervised semantic segmentation. OVDiff~\cite{ovdiff} applies SD to generate supportive images and encodes them into prototypes for open-vocabulary segmentation (OVS). In WSSS, DiG~\cite{DiG} leverages SD to enhance the locality of Vision Transformer, while Attn2mask~\cite{attn2mask} and ICAD~\cite{ICAD} utilize it for data augmentation. Unlike previous WSSS methods that require complex training processes to learn from diffusion models, our DiCLIP efficiently transfers the dense knowledge from SD in a training-free manner. In addition, instead of generating prototypes as OVDiff, we explore a cache model that stores both generated image features and their class labels as key-value pairs. It not only facilitates effective patch-wise knowledge retrieval but also supports a dynamic adapter for CAM generation, demonstrating superior performance in WSSS.
\vspace{-1em}
\subsection{Key-value Cache Model}
The cache model operates similarly to the attention mechanism, where training image features and their labels serve as key-value pairs, and the test samples act as queries, enabling knowledge retrieval during inference adaptation~\cite{ViT}. Once the cache is built, the retrieval process acts in a training-free manner and no trainable parameters are required. It has received increasing interest across various fields. For text generation, PSMM~\cite{PSMM} and Unbounded Cache~\cite{unboundcachemodel} build a cache model to store massive training samples to capture long-term dependencies. For few-shot classification, TiP-Adapter~\cite{tipadapter} caches labeled images to achieve predictions for unlabeled test samples with image-wise knowledge retrieval. \hl{Different from TiP-Adapter~{\cite{tipadapter}}, we pioneer a cache model for CAM generation via a novel patch-wise knowledge retrieval. To explore the patch-level predictions, our method highlights the following perspectives: (1) Both foreground and background cache models are built to reduce noise during the dense retrieval; (2) Text embeddings are used to capture intra- and inter-class differences among cached Keys, leading to more reliable knowledge processing; (3) Learnable class-agnostic key-value prompts are incorporated and further initiate a learnable adapter for dynamic CAM generation in WSSS.}

\begin{figure*}[h!]
  \centering
  \includegraphics[width=18.2cm]{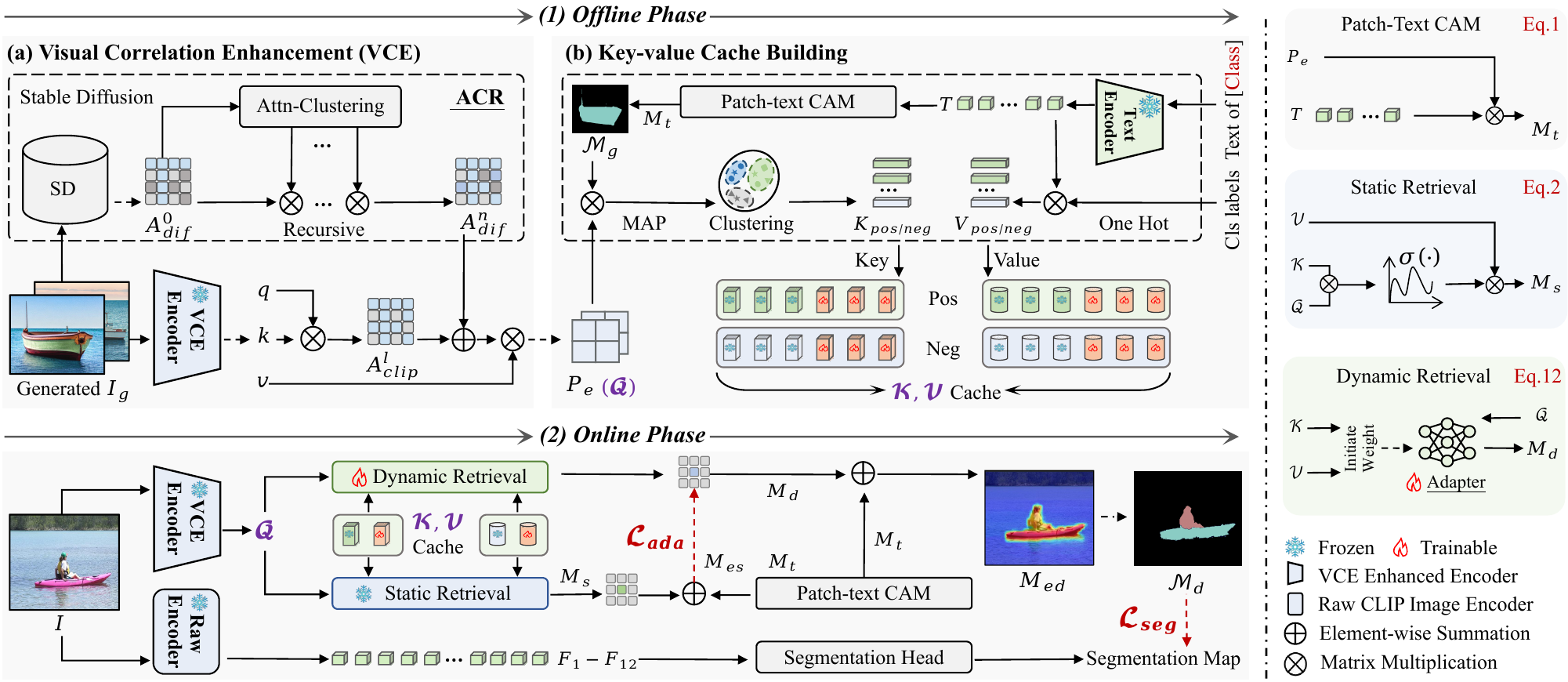}
    \caption{{Framework of DiCLIP. Our method leverages Stable Diffusion (SD) to enhance CLIP's dense capability through offline and online phases. (1)} \textbf{{Offline phase}}{: (a) We propose Visual Correlation Enhancement (VCE) to enrich CLIP's image encoder, forming the VCE encoder. The input image (e.g., generated $I_{g}$) is processed by SD to extract self-attention maps, which are recursively refined by our Attention Clustering Refinement (ACR). The refined attention $A_{dif}^{n}$ diversifies CLIP’s smooth attention $A_{clip}^{l}$, producing diverse features $P_e$. (b) With $P_e$ and text embedding $T$, patch-text CAMs $M_t$ are generated and refined into pseudo labels $\mathcal{M}_{g}$. Keys are obtained by masking $P_e$ with $\mathcal{M}_{g}$, and Values are one-hot class labels, forming a non-trainable key–value cache. (2)} \textbf{{Online phase}}{: The VCE-enhanced encoder and the cache jointly improve CLIP's dense knowledge in a training-efficient manner. VCE diversifies image features, while the cache enhances text–image alignment through static and dynamic dense retrieval, reframing patch-text CAMs into enhanced CAMs $M_{ed}$. Finally, features from raw CLIP image encoder are fed into the segmentation head, with $\mathcal{M}_{d}$ refined from $M_{ed}$ as supervision. See Section{~\ref{sec.3.2}} for details.}
}
   \label{fig.2}
  \vspace{-1em}
\end{figure*}

\section{Methodology}
In this section, the preliminaries for patch-text aligned CAM and dense knowledge retrieval are introduced in Section \ref{sec.3.1}. Framework overview of DiCLIP is presented in Section \ref{sec.3.2}. The proposed Visual Correlation Enhancement (VCE) and Text Semantic Augmentation (TSA) modules are specifically introduced in Section \ref{sec.3.3} and Section \ref{sec.3.4}, respectively. The training objectives for DiCLIP are detailed in Section \ref{sec.3.5}. 
\vspace{-.5em}
\subsection{Preliminaries}
\label{sec.3.1}

\hl{\textbf{A.1) Problem Formulation.} Given the input image $I$ and a class label space $\{1, \dots, C\}$, our goal is to generate CAMs that serve as pseudo-labels for WSSS. The observation model is based on a frozen CLIP backbone, which extracts visual patch features
$P \in \mathbb{R}^{H \times W \times D}$ and class-specific text embeddings
$T \in \mathbb{R}^{D \times C}$, where $H$ and $W$ are spatial sizes, $C$ is the number of classes, and $D$ is the feature dimension.

The main optimization objective is to obtain high-quality CAMs by measuring patch-wise similarity between $P$ and $T$, under the constraint that only image-level labels are available. However, due to the image-text paired training paradigm, CLIP focuses on global representations, leaving the dense knowledge across visual and text modalities limited. As a result, CAMs from patch-text alignment are still suboptimal.

To overcome this, DiCLIP enhances CLIP's dense knowledge by harnessing the reliable spatial consistency and generative power of diffusion model, built upon two key assumptions:
(1) \emph{Spatial Consistency}: attention maps from SD preserve coherent spatial structures and serve as reliable spatial priors to enhance CLIP's spatial awareness; (2) \emph{Generative Validity}: images generated by SD contain reliable prototypical features, which help improve the patch-text alignment of CLIP.
By integrating these priors, DiCLIP enables better patch-text aligned CAMs and improves downstream segmentation performance.}

{\bf A.2) Patch-text Aligned CAM.} Unlike previous WSSS methods that generate CAMs via image-text matching, our DiCLIP adopts a patch-text alignment paradigm. Specifically, we first send image $I$ and the corresponding text templates into CLIP, acquiring the visual features $P$ and text embeddings $T$. Then the CAMs $M_t$ is generate by calculating the patch-wise cosine similarities between $P$ and $T$:
\begin{equation}                            
    {M_{t}=\operatorname{Norm}\left(\operatorname{cosim}\left({P}, {T}\right)\right).}
    \label{eq:1}
\end{equation}
where $M_{t} \in \mathbb{R}^{H \times W \times C}$, $\operatorname{Norm}(\cdot)$ is the min-max normalization, and $\operatorname{cosim}(\cdot)$ calculates the cosine similarity . 

{\bf A.3) Dense Knowledge Retrieval.} The cache model stores the training features and the corresponding one-hot class labels as key-value pairs. Similar to the attention mechanism~\cite{ViT}, the test samples act as queries, enabling a training-free knowledge retrieval during inference. Tip-Adapter~\cite{tipadapter} has introduced it for few-shot classification via an image-wise knowledge retrieval. In this work, we pioneer a cache model for CAM generation via a novel patch-wise knowledge retrieval. Once the Key $\mathcal{K} \in \mathbb{R}^{U \times D}$ and Value $\mathcal{V} \in \mathbb{R}^{U \times C}$ are cached (where $U$ is the number of key-value pairs), we view each patch of visual features $P \in \mathbb{R}^{H \times W \times D}$ as a query $\mathcal{Q}$, retrieving knowledge from Keys and Values of cache model. The dense knowledge retrieval operation is denoted as:
\begin{equation}                            
    \operatorname{RE}({\mathcal{Q}},\mathcal{K},\mathcal{V})=\operatorname{Norm}(\sigma({\mathcal{Q}} \mathcal{K}^{{T}}) \mathcal{V}),
    \label{eq:2}
\end{equation}
where $\sigma(\cdot)$ is the activation function. With this operation, we generate cache-based CAM and use it to enhance patch-text aligned CAM $M_{t}$, leading to superior performance for WSSS. More Details are specified in Section \ref{sec.3.4}.

\subsection{{Framework Overview}}
\label{sec.3.2}

{DiCLIP leverages Stable Diffusion (SD) to enhance CLIP's dense knowledge through offline and online phases, reframing patch–text CAM into a visual knowledge retrieval paradigm.

In the offline phase, we maintain a Key-Value cache in a training-free manner, as illustrated in Fig.{~\ref{fig.2}} (1).}

\textbf{{(I) Enhancing CLIP’s Image Encoder with VCE.}} {In Fig.{~\ref{fig.2}} (a), given an input image (generated image $I_g$ or real image $I$), we first pass it through SD to extract self-attention maps. These maps are then recursively refined by our Attention Clustering Refinement (ACR) module to obtain $A_{dif}^{n}$. Next, $A_{dif}^{n}$ is injected into CLIP by diversifying the original smooth attention maps $A_{clip}^{l}$ from the last $L$ layers. This non-parametric calibration enhances CLIP’s image encoder, forming the VCE encoder. It produces enriched visual features $P_e$. Using $P_e$ with text embeddings $T$, patch-text CAMs $M_t$ are generated and further refined into the pseudo labels $\mathcal{M}_{g}$.}

\textbf{{(II) Building a Key–value Cache Model.}} {In Fig.{~\ref{fig.2}} (b), given $P_e$ and pseudo labels $\mathcal{M}_{g}$, we construct a cache in a training-free manner. Specifically, the Keys $\mathcal{K}$ are obtained by masking $P_e$ with $\mathcal{M}_{g}$ with a clustering operation, while the Values $\mathcal{V}$ correspond to one-hot class labels, weighted by text embeddings $T$. This cache encodes SD's prior knowledge and serves as an external memory for dense knowledge retrieval.

In the online phase, we achieve WSSS in a training-efficient manner, as illustrated in Fig.{~\ref{fig.2}} (2).}

\textbf{{(III) Dense Knowledge Retrieval.}} {With the VCE-enhanced encoder and the key–value cache, we first view $P_{e}$ from input image $I$ as query and generate the static cache CAM $M_s$ with static knowledge retrieval in Eq.{~\ref{eq:2}}. We further conduct dynamic knowledge retrieval by initiating a dynamic adapter with the cache model, generating dynamic cache CAM $M_d$ via Eq.{~\ref{eq:12}}. We generate $M_{es}$ by adding $M_s$ to $M_t$ and refine it to supervise $M_d$ with loss $\mathcal{L}_{ada}$.}

\textbf{{(IV) Segmentation Training.}} The CAMs $M_{ed}$ are obtained by enhancing $M_{t}$ with $M_{d}$, and subsequently refined into the final pseudo labels $\mathcal{M}_{d}$. \hl{Following{~\cite{18}}, we extract image features from the raw CLIP image encoder and feed them into the segmentation head. Then $\mathcal{M}_{d}$ is employed as supervision to train the segmentation head in an end-to-end manner.}

\subsection{Visual Correlation Enhancement}
\label{sec.3.3}
\textbf{C.1) Suboptimal Self-attention of CLIP.} Due to CLIP's image-text pairing nature, its visual features lack spatial-awarded information, leading to suboptimal localization maps obtained through patch-text alignment. To investigate it, let us review the self-attention mechanism first. Given the input image $I$, features $P_l$ from $l$-th layer of CLIP $\phi_{CLIP}(\cdot)$ are generated. It is projected into three different spaces $\{q,k,v\}$, named query, key, and value, respectively. The attention map is generated by calculating the similarity between $q$ and $k$ as:
\begin{equation}                            
    A_{clip}^{l}=\operatorname{softmax}\left({q}^{{T}} {k} / \sqrt{{D}_{{k}}}\right),
    \label{eq:3}
\end{equation}
where $D_k$ is the channels of $k$. Then the similarity weights in attention maps guide the aggregation of tokens of $v$ to generate the output features $P_{l+1}$. 

\begin{figure}[t]
  \centering
  \includegraphics[width=8.7cm]{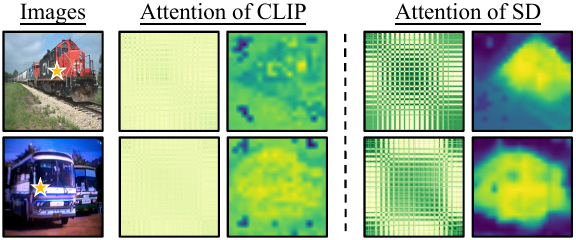}
   \caption{Attention maps from CLIP and Stable Diffusion (SD). For each model, the affinity map (left) and the queried attention response (right) given the query patch (yellow star) are showcased. CLIP's attention is smoothing while SD generates more distinctive and spatially aware attention maps.}
   \label{fig.3}
   \vspace{-.5em}
\end{figure}

However, as shown in Fig.~\ref{fig.3}, CLIP produces overly smooth attention maps, which lack sufficient spatial detail for dense prediction.
On the other hand, the SD model uses a transformer-based UNet to recover images from noisy data, containing rich spatial contexts within the self-attention maps. Inspired by it, we utilize diffusion's attention maps to enrich the spatial information of CLIP's attention layer, providing diverse visual features for CAM generation.

\begin{figure}[t]
  \centering
  \includegraphics[width=8.7cm]{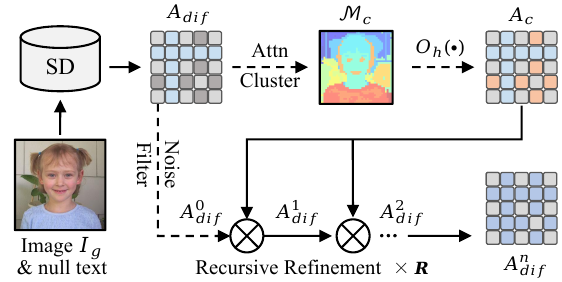}
   \caption{Illustration of Attention Clustering Refinement (ACR) module. Since the extracted attention $A_{dif}$ from Stable Diffusion (SD) holds noisy correlations, we first generate a clustering mask $\mathcal{M}_{c}$ to capture the reliable patch-wise relations $A_{c}$ and use it to recursively refine the attention.}
   \label{fig.4}
   \vspace{-.5em}
\end{figure}
\textbf{C.2) Attention Clustering Refinement.} To further improve the reliability of extracted attention maps from diffusion, we propose the Attention Clustering Refinement (ACR) before calibrating CLIP. \hl{As illustrated in Fig.~{\ref{fig.4}}, we first send the input image $I$ (or generated image $I_g$) to SD $\phi_{SD}(\cdot)$ along with a null textual prompt. Then we extract the self-attention maps $A_{dif} \in \mathbb{R}^{HW \times HW}$ from the denoised UNet following~{\cite{diffsegmenter}}}. As the attention maps reflect the spatial similarities among different regions, we leverage the online clustering algorithm~\cite{kl_clustering} to cluster the attention maps into different semantic groups:
\begin{equation}                            
    \mathcal{M}_{{c}}=\operatorname{Clustering}({A}_{{dif}}, {B}),
    \label{eq:4}
\end{equation}
where $B$ is the number of clustering centroids, and $\mathcal{M}_c \in \mathbb{R}^{H \times W}$ is the generated attention masks indicating different semantics of groups. \hl{Then $\mathcal{M}_c$ is converted to the patch-wise affinity label $A_c \in \mathbb{R}^{HW \times HW}$. It represents the patch-wise relations based on the clustering operation:}
\begin{equation}                            
    {A}_{{c}}={O}_{{h}}\left(\mathcal{M}_{{c}}\right)^{{T}} {O}_{{h}}\left(\mathcal{M}_{{c}}\right),
    \label{eq:5}
\end{equation}
where ${O}_{{h}}(\cdot)$ is the one-hot encoding function and ${O}_{{h}}(\mathcal{M}_{{c}}) \in \mathbb{R}^{C \times HW}$. To propagate the most reliable relations of $A_{dif}$, we first adopt a threshold $\epsilon$ to filter the noisy relations in low values and generate the reliable attention $A_{dif}^{0}$. Then $A_c$ is applied to recursively refine $A_{dif}^{0}$, further promoting the confident correlations within the same semantics while suppressing the irrelevant. The refinement is denoted as:
\begin{equation}                            
    {A}_{{dif}}^{{n}}={A}_{{c}} \times {A}_{{dif}}^{{n}-1}, {n}=1, \ldots, {R},
    \label{eq:6}
\end{equation}
where $A_{dif}^{0}=A_{dif}*\mathbb{1}(A_{dif}>\epsilon)$ and $\mathbb{1}(\cdot)$ is the indicator function. $R$ represents the number of iterations.

{An additive fusion strategy is adopted to combine the attention maps from CLIP and SD. By adding SD attention to CLIP’s, we enhance spatial diversity while preserving CLIP’s inherent image–text alignment. This design is analogous to the principle of LoRA~{\cite{lora}}, where a trainable low-rank update is directly added to the pretrained weight matrix for task adaptation. Similarly, we apply SD attention as an off-the-shelf beneficial bias without introducing additional trainable parameters, thus enabling a training-free mode of CAM generation.} Specifically, the refined $A_{dif}^{n}$ is added as a distributional shift to CLIP’s original attention maps $A_{clip}^{l}$, where the last $L$ layers are calibrated, progressively propagating the spatial knowledge. The enhanced attention $A_{en}^{l}$ is denoted as:
\begin{equation}                            
    {A}_{{en}}^{{l}}={A}_{{clip}}^{{l}} + \alpha {A}_{{dif}}^{{n}},
    \label{eq:7}
\end{equation}
where $\alpha$ is the weight to balance the contribution from $A_{dif}^{n}$ and is set to $1.0$ by default. Finally, we form the VCE encoder and extract the enhanced visual features $P_{e}$ from the last enhanced attention layer, generating patch-text aligned CAM $M_{t}$ following Eq.~\ref{eq:1}. With this non-parametric enhancement, our VCE generates $M_{t}$ in a training-free manner, even comparable to most training-required WSSS methods.

\subsection{Text Semantic Augmentation}
\label{sec.3.4}
\textbf{D.1) Key-value Cache Building.} We leverage SD's generative power to maintain a key-value cache model, further enhancing CAM via the novel patch-wise knowledge retrieval, as shown in Fig~\ref{fig.2} (b). Our cache model highlights the dual branches for both foregrounds and backgrounds, effectively alleviating the adverse impact of the dense retrieval. 

To generate the Keys in the cache model, we first apply SD to generate single-class images $I_g$ \hl{(maintaining the semantic purity and avoiding the co-occurrence issue~{\cite{seco}})}. They are sent to VCE-enhanced image encoder $\phi(\cdot)_{VCE}$ to generate corresponding pseudo masks $\mathcal{M}_g$ via Eq.~\ref{eq:1}. With the image-mask pairs, we use mask average pooling (MAP) to summarize the foreground semantics $F_{pos}$:
\begin{equation}                            
    {F}_{{pos}}=\frac{\sum \phi_{VCE}\left({I}_{{g}}\right) * \mathbb{1}\left(\mathcal{M}_{{g}}>0\right)}{\sum \mathbb{1}\left(\mathcal{M}_{{g}}>0\right)}.
    \label{eq:8}
\end{equation}
Importantly, background semantics $F_{neg}$ is also generated with the same operation, where $\mathbb{1}(\mathcal{M}_{{g}}=0)$ at this time.

\hl{To select the most representative cases, Kmeans~{\cite{kmeans}} is used to cluster the class features $F_{pos}^{y}$ within each class $y$. The clustering centroids are stored as Keys.} The foreground Keys ${K}_{{pos}}\in \mathbb{R}^{(C \times E) \times D}$ are denoted as:
\begin{equation}                            
    {K}_{{pos}}=\operatorname{Concat}\left[\operatorname{Kmeans}\left({F}_{{pos}}^{{y}}, {E}\right)\right]_{{y}=1}^{{C}},
    \label{eq:9}
\end{equation}
where $\operatorname{Concat}[\cdot]$ concatenates samples along batch dimension and $E$ is the number of centroids. The negative Keys ${K}_{{neg}} \in \mathbb{R}^{E \times D}$ are generated by clustering $F_{neg}$ as well. 

To build the Values in the cache model, we apply text embeddings $T$ to assign weights to one-hot class labels, capturing the intra- and inter-class differences among cached Keys. Specifically, given $K_{pos}$, its one-hot class labels are denoted as: $\mathcal{Y} \in \mathbb{R}^{(C \times E) \times \mathcal{C}}$, where $C \times E$ is the total number of Values and $\mathcal{C}=C+1$ means we take the background as a class. We calculate the similarities between $T$ and $K_{pos}$. Then the similarity scores are assigned to $\mathcal{Y}$, measuring the importance of each instance in $K_{pos}$. The foreground Values $V_{pos} \in \mathbb{R}^{(C \times E) \times \mathcal{C}}$ are formulated as:
\begin{equation}                            
    {V}_{{pos}}=\mathcal{Y} * \operatorname{softmax}(K_{pos}T).
    \label{eq:10}
\end{equation}
For background Values $V_{neg}$, we directly use their one-hot label considering the various background semantics within the images. With positive and negative Keys and Values, we concatenate them along the batch dimension to form the static Keys and Values as: $\mathcal{K}_s \in \mathbb{R}^{(\mathcal{C} \times E) \times D}$ and $\mathcal{V}_s \in \mathbb{R}^{(\mathcal{C} \times E) \times \mathcal{C}}$. 

\textbf{D.2) Static Knowledge Retrieval.} With $\mathcal{K}_s$, $\mathcal{V}_s$, and $P_e$, we generate CAM via Eq.~\ref{eq:2}. To reliably excluding noise during the static dense retrieval, we further divide it into foreground part $M_{s}^{pos} \in \mathbb{R}^{(H \times W) \times C}$ and background part $M_{s}^{neg} \in \mathbb{R}^{(H \times W) \times 1}$ along the class dimension. We use $M_{s}^{neg}$ to refine $M_{s}^{pos}$ by multiplying them, generating the final static cache CAM $M_{s}$. Then $M_s$ is used to enhance the patch-text aligned $M_t$. The enhanced static CAM $M_{es}$ is generated as: 
\begin{equation}                            
    M_{es}=M_{t} + \beta M_{s}^{pos}*M_{s}^{neg},
    \label{eq:11}
\end{equation}
where $\beta$ is the weight to balance the contribution. 

\textbf{D.3) Dynamic Knowledge Retrieval.} Although DiCLIP is competent to generate high-quality CAM $M_{es}$ even without any training, {its performance is potentially hindered by the fixed features or synthetic bias from generated cache model.} To further unleash the potential of cache model, we build a learnable adapter for dynamic knowledge retrieval. Specifically, $N$ learnable Key and Value prompts are first incorporated into $[\mathcal{K}_s, \mathcal{V}_s]$. The prompt-incorporated Keys and Values are now denoted as: $\mathcal{K}_d \in \mathbb{R}^{(\mathcal{C} \times E + N) \times D}$ and $\mathcal{V}_d \in \mathbb{R}^{(\mathcal{C} \times E + N) \times \mathcal{C}}$. \hl{Inspired by~{\cite{tipadapter}}, we view the dense retrieval process in Eq.~{\ref{eq:2}} as a form of a two-layer MLP adapter. The cached Keys and Values are used to initialize the linear weights of each layer for efficient knowledge retrieval. Consequently, it transforms static visual knowledge retrieval into a dynamic process for CAM generation.} The dynamic cache CAM is formulated as: 

\begin{equation}                            
    {M}_{{d}}=\sigma\left({P}_{{e}}{W}_1^{{T}}+{b}_1\right) {W}_2+{b}_2,
    \label{eq:12}
\end{equation}
where $W_1$ and $W_2$ are initiated with $\mathcal{K}_d$ and $\mathcal{V}_d$, $b_1$ and $b_2$ are initiated to 0. With $M_{d}$, we extract its foreground part $M_{d}^{pos} \in \mathbb{R}^{(H \times W) \times C}$ to enhance the patch-text aligned CAM. The final enhanced dynamic CAM $M_{ed}$ is generated as: 
\begin{equation}                            
   M_{ed}=M_{t} + \beta M_{d}^{pos}.
    \label{eq:13}
\end{equation}
Then $M_{ed}$ is subsequently refined into pseudo masks $\mathcal{M}_{d}$ with PAR~\cite{12}, which supervise the segmentation decoder. 

\subsection{Training Objectives}
\label{sec.3.5}
To supervise the dynamic adapter, {we refine the enhanced static CAM $M_{es}$ to pseudo labels $\mathcal{M}_s$ and use it as supervision with cross entropy loss:}
\begin{equation}                            
    \mathcal{L}_{ada} = CE(M_{d}, \mathcal{M}_s).
    \label{eq:14}
\end{equation}

It is noted that once the cache is built from the diffusion model, DiCLIP is trained in a single-stage paradigm. {We use the dynamic pseudo masks $\mathcal{M}_{d}$ to train a lightweight Transformer-based segmentation head, while extracting frozen features $\{F_1,..., F_{12}\}$ from the original CLIP image encoder for segmentation, following{~\cite{18}}. This design improves the feasibility of DiCLIP during the segmentation inference stage by eliminating the dependence on SD.} Cross entropy loss $\mathcal{L}_{seg}$ is used for segmentation. Our total loss objective is:
\begin{equation}                           
\mathcal{L}_{\text {DiCLIP}}=\mathcal{L}_{{seg}}+\gamma \mathcal{L}_{\text {ada}},
    \label{eq:15}
\end{equation}
where $\gamma$ is a loss weight. By only training the adapter and a segmentation head, DiCLIP achieves strong WSSS performance and significantly reduces the training cost.

\section{Experiments}
\subsection{Main Results}
{\bf Performance of Semantic Segmentation.}~Table~\ref{tab.1} and ~Table~\ref{tab.2} quantitatively report the segmentation comparisons between our DiCLIP and other recent state-of-the-art (SOTA) methods on PASCAL VOC 2012~\cite{34} and MS COCO 2014~\cite{35}. For VOC dataset, ours achieves $78.8 \%$ mIoU on val set and $78.9 \%$ mIoU on test set. Compared to other CLIP-based counterparts, such as the single-stage SOTA WeCLIP~\cite{18} and the multi-stage SOTA WeakCLIP~\cite{37}, DiCLIP outperforms them by $2.4 \%$ and $3.7 \%$ on val set. For COCO dataset, Table~\ref{tab.2} shows that DiCLIP achieves $48.7\%$ mIoU on val set, while WeCLIP and POT~\cite{POT} only achieve $47.1\%$ and $47.9\%$. Noticeably, even without post-processing CRF~\cite{CRF}, our method still shows competitive results on both datasets. 
\newcommand{\tablestyle}[2]{\setlength{\tabcolsep}{#1}\renewcommand{\arraystretch}{#2}\centering\footnotesize}
\begin{table}[!t]
\centering
\caption{Segmentation comparisons on VOC. $Sup.$: supervision. $\mathcal{I}$: image-level labels. $\mathcal{SA}$: saliency maps. $\mathcal{L}$: Language.}
    \tablestyle{7.pt}{1.084}
    \scalebox{1.}
    {
    \footnotesize
    \begin{tabular}{l|cccccc}
    \toprule
    \multicolumn{1}{l|}{\multirow{2}{*}{Method}} & \multicolumn{1}{c|}{\multirow{2}{*}{Sup.}} & \multicolumn{1}{c|}{\multirow{2}{*}{Backbone}} & \multicolumn{2}{c}{VOC}                                   \\ \cmidrule{4-5} 
    \multicolumn{1}{l|}{}                        & \multicolumn{1}{c|}{}    & \multicolumn{1}{c|}{}                   & Val           & Test                    \\ \midrule                     

    \multicolumn{1}{l|}{L2G~\cite{L2G} \tiny CVPR'2022}                                             & \multicolumn{1}{c|}{$\mathcal{I}+\mathcal{SA}$}                                  & \multicolumn{1}{c|}{RN101}                                  & 72.1          & \multicolumn{1}{c}{71.7}                                           \\

    \multicolumn{1}{l|}{RCA~\cite{RCA} \tiny CVPR'2023}                                             & \multicolumn{1}{c|}{$\mathcal{I}+\mathcal{SA}$}                                  & \multicolumn{1}{c|}{RN38}                                   & 72.2          & \multicolumn{1}{c}{72.8}                                           \\

    \multicolumn{1}{l|}{Mat-label\cite{matlabel} \tiny ICCV'2023}                                             & \multicolumn{1}{c|}{$\mathcal{I}+\mathcal{SA}$}                                  & \multicolumn{1}{c|}{RN101}                                   & 73.3          & \multicolumn{1}{c}{74.0}                                            \\

    \multicolumn{1}{l|}{\textcolor{black}{GSM\cite{GSM} \tiny AAAI'2021}}                                             & \multicolumn{1}{c|}{$\mathcal{I}$}                                  & \multicolumn{1}{c|}{RN101}                                   & \textcolor{black}{68.2}          & \multicolumn{1}{c}{\textcolor{black}{68.5}}                                            \\
    
    \multicolumn{1}{l|}{AFA~\cite{12} \tiny CVPR'2022}                                             & \multicolumn{1}{c|}{$\mathcal{I}$}                                    & \multicolumn{1}{c|}{MiT-B1}                                 & 66.0          & \multicolumn{1}{c}{66.3}                                            \\

    \multicolumn{1}{l|}{W-OoD\cite{wood} \tiny CVPR'2022}                                             & \multicolumn{1}{c|}{$\mathcal{I}$}                                  & \multicolumn{1}{c|}{RN101}                                   & 69.8          & \multicolumn{1}{c}{69.9}                                            \\

    \multicolumn{1}{l|}{FPR\cite{fpr} \tiny ICCV'2023}                                             & \multicolumn{1}{c|}{$\mathcal{I}$}                                  & \multicolumn{1}{c|}{RN101}                                   & 70.0          & \multicolumn{1}{c}{70.6}                                            \\

    

    \multicolumn{1}{l|}{BECO~\cite{BECO} \tiny CVPR'2023}                                            & \multicolumn{1}{c|}{$\mathcal{I}$}                                    & \multicolumn{1}{c|}{RN101}                                  & 73.7          & \multicolumn{1}{c}{73.5}                                            \\
    
    \multicolumn{1}{l|}{DiG~\cite{DiG} \tiny ECCV'2024}                                             & \multicolumn{1}{c|}{$\mathcal{I}$}                                    & \multicolumn{1}{c|}{RN101}                                  & 73.9          & \multicolumn{1}{c}{73.7}                                            \\
    \multicolumn{1}{l|}{PLDA~\cite{PLDA} \tiny TIP'2024}                                     & \multicolumn{1}{c|}{$\mathcal{I}$}                                    & \multicolumn{1}{c|}{RN101}                                                                     & 69.7    & \multicolumn{1}{c}{71.7}      \\
    \multicolumn{1}{l|}{SSC~\cite{t1} \tiny TIP'2024}                                     & \multicolumn{1}{c|}{$\mathcal{I}$}                                    & \multicolumn{1}{c|}{RN101}                                                                     & 72.7    & \multicolumn{1}{c}{72.8}     \\
    \multicolumn{1}{l|}{DGRM\cite{t1} \tiny TCSVT'2025}                                             & \multicolumn{1}{c|}{$\mathcal{I}$}                                  & \multicolumn{1}{c|}{RN101}                                   & 71.4          & \multicolumn{1}{c}{71.3}                                            \\
    \multicolumn{1}{l|}{MCTformer+~\cite{33} \tiny TPAMI'2024}                                     & \multicolumn{1}{c|}{$\mathcal{I}$}                                    & \multicolumn{1}{c|}{RN38}                                   & 74.0          & \multicolumn{1}{c}{73.6}                                            \\

    \multicolumn{1}{l|}{CLIMS~\cite{15} \tiny CVPR'2022}                                           & \multicolumn{1}{c|}{$\mathcal{I}+\mathcal{L}$}                                  & \multicolumn{1}{c|}{RN101}                                  & 70.4          & \multicolumn{1}{c}{70.0}                                               \\
    
    \multicolumn{1}{l|}{CLIP-ES~\cite{16} \tiny CVPR'2023}                                         & \multicolumn{1}{c|}{$\mathcal{I}+\mathcal{L}$}                                  & \multicolumn{1}{c|}{RN101}                                  & 72.2          & \multicolumn{1}{c}{72.8}                                            \\
    
    \multicolumn{1}{l|}{PSDPM~\cite{PSDPM} \tiny CVPR'2024}                                           & \multicolumn{1}{c|}{$\mathcal{I}+\mathcal{L}$}                                  & \multicolumn{1}{c|}{RN101}                                  & 74.1          & \multicolumn{1}{c}{74.9}                                          \\
    \multicolumn{1}{l|}{CPAL~\cite{37} \tiny CVPR'2024}                                            & \multicolumn{1}{c|}{$\mathcal{I}+\mathcal{L}$}                                  & \multicolumn{1}{c|}{RN101}                                  & 74.5          & \multicolumn{1}{c}{74.7}                                            \\ 

    \multicolumn{1}{l|}{ToCo~\cite{22} \tiny CVPR'2023}                                            & \multicolumn{1}{c|}{$\mathcal{I}$}                                    & \multicolumn{1}{c|}{ViT-B}                                  & 71.1          & \multicolumn{1}{c}{72.2}                                            \\
    \multicolumn{1}{l|}{DuPL~\cite{DuPL} \tiny CVPR'2024}                                            & \multicolumn{1}{c|}{$\mathcal{I}$}                                    & \multicolumn{1}{c|}{ViT-B}                                  & 73.3          & \multicolumn{1}{c}{72.8}                                            \\
    \multicolumn{1}{l|}{SeCo~\cite{seco} \tiny CVPR'2024}                                            & \multicolumn{1}{c|}{$\mathcal{I}$}                                    & \multicolumn{1}{c|}{ViT-B}                                  & 74.0          & \multicolumn{1}{c}{73.8}                                            \\
    \multicolumn{1}{l|}{\textcolor{black}{FFR~\cite{FFR} \tiny CVPR'2025}}                                            & \multicolumn{1}{c|}{$\mathcal{I}$}                                    & \multicolumn{1}{c|}{ViT-B}                                  & \textcolor{black}{76.0}          & \multicolumn{1}{c}{\textcolor{black}{75.5}}                                            \\
    \multicolumn{1}{l|}{\textcolor{black}{PCRE~\cite{PCRE} \tiny CVPR'2025}}                                            & \multicolumn{1}{c|}{$\mathcal{I}$}                                    & \multicolumn{1}{c|}{ViT-B}                                  & \textcolor{black}{75.5}          & \multicolumn{1}{c}{\textcolor{black}{75.9}}                                            \\
    \multicolumn{1}{l|}{\textcolor{black}{MuP-VSS~\cite{MuP-VSS} \tiny CVPR'2025}}                                            & \multicolumn{1}{c|}{$\mathcal{I}$}                                    & \multicolumn{1}{c|}{ViT-B}                                  & \textcolor{black}{73.6}          & \multicolumn{1}{c}{\textcolor{black}{74.7}}                                            \\
    \multicolumn{1}{l|}{\textcolor{black}{ToMA~\cite{ToMa} \tiny ICCV'2025}}                                            & \multicolumn{1}{c|}{$\mathcal{I}$}                                    & \multicolumn{1}{c|}{ViT-B}                                  & \textcolor{black}{72.7}          & \multicolumn{1}{c}{\textcolor{black}{73.5}}                                            \\
    \multicolumn{1}{l|}{\textcolor{black}{MoRe~\cite{MoreZW} \tiny AAAI'2025}}                                            & \multicolumn{1}{c|}{$\mathcal{I}$}                                    & \multicolumn{1}{c|}{ViT-B}                                  & \textcolor{black}{76.4}          & \multicolumn{1}{c}{\textcolor{black}{75.0}}                                            \\
    \multicolumn{1}{l|}{DIAL~\cite{DIAL} \tiny ECCV'2024}                                            & \multicolumn{1}{c|}{$\mathcal{I}+\mathcal{L}$}                                  & \multicolumn{1}{c|}{ViT-B}                                  & 74.5          & \multicolumn{1}{c}{74.9}                                            \\
    \multicolumn{1}{l|}{WeakCLIP~\cite{weakclip} \tiny IJCV'2024}                                            & \multicolumn{1}{c|}{$\mathcal{I}+\mathcal{L}$}                                     & \multicolumn{1}{c|}{ViT-B}                                  & 75.1          & \multicolumn{1}{c}{74.9}                                            \\
    \multicolumn{1}{l|}{WeCLIP~\cite{18} \tiny CVPR'2024}                                          & \multicolumn{1}{c|}{$\mathcal{I}+\mathcal{L}$}                                  & \multicolumn{1}{c|}{ViT-B}                                  & 76.4          & \multicolumn{1}{c}{77.2}                                           \\
    \multicolumn{1}{l|}{\textcolor{black}{POT~\cite{POT} \tiny CVPR'2025}}                                          & \multicolumn{1}{c|}{$\mathcal{I}+\mathcal{L}$}                                  & \multicolumn{1}{c|}{ViT-B}                                  & \textcolor{black}{76.1}          & \multicolumn{1}{c}{\textcolor{black}{76.7}}                                           \\
    \rowcolor[HTML]{EFEFEF} 
    \multicolumn{1}{l|}{\cellcolor[HTML]{EFEFEF}\textbf{DiCLIP (w/o CRF)}} & \multicolumn{1}{c|}{\cellcolor[HTML]{EFEFEF}\textbf{$\mathcal{I}+\mathcal{L}$}} & \multicolumn{1}{c|}{\cellcolor[HTML]{EFEFEF}\textbf{ViT-B}} & \textbf{77.1} & \multicolumn{1}{c}{\cellcolor[HTML]{EFEFEF}\textbf{77.3}}  \\
    \rowcolor[HTML]{EFEFEF} 
    \multicolumn{1}{l|}{\cellcolor[HTML]{EFEFEF}\textbf{DiCLIP (Ours)}}          & \multicolumn{1}{c|}{\cellcolor[HTML]{EFEFEF}\textbf{$\mathcal{I}+\mathcal{L}$}} & \multicolumn{1}{c|}{\cellcolor[HTML]{EFEFEF}\textbf{ViT-B}} & \textbf{78.8} & \multicolumn{1}{c}{\cellcolor[HTML]{EFEFEF}\textbf{78.9}}  \\ \bottomrule
    \end{tabular}
    }
    \label{tab.1}
   \vspace{-1.5em}
\end{table}

\begin{figure*}
  \centering
    \includegraphics[width=18.1cm]{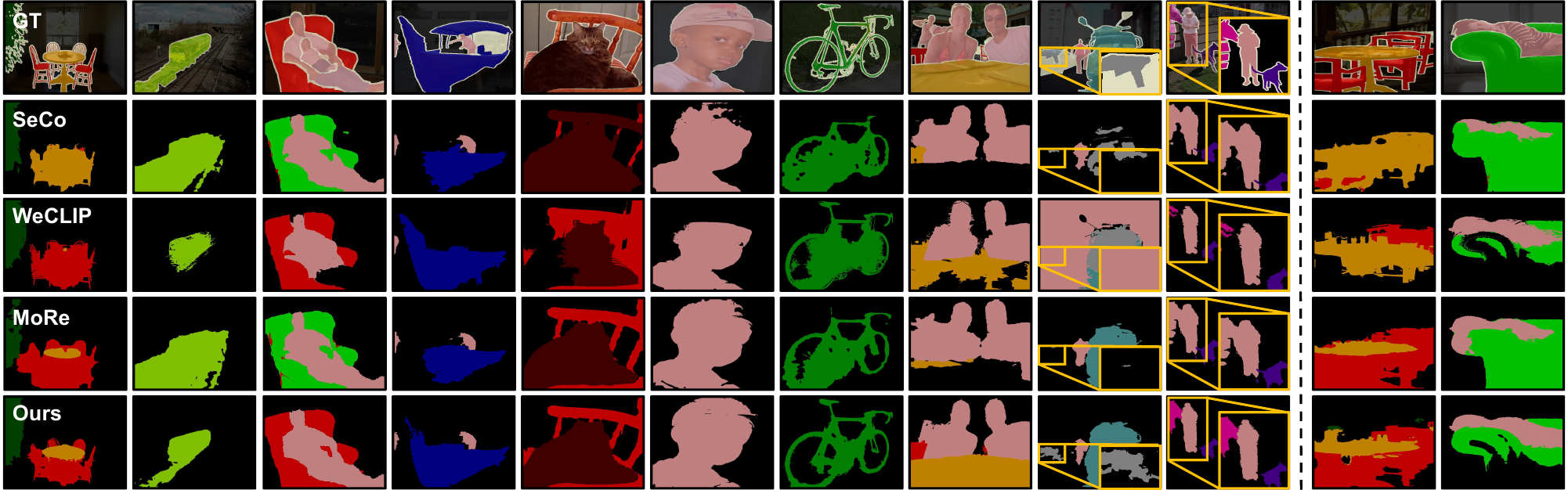}
  \caption{\hl{Qualitative segmentation comparisons on PASCAL VOC 2012. We compare DiCLIP with the recent CLIP-based state-of-the-art method WeCLIP~{\cite{weakclip}} and two other advanced approaches, SeCo~{\cite{seco}} and MoRe~{\cite{MoreZW}}. Small and off-center objects are highlighted with yellow rectangles, and failure cases are shown in the last two columns. It shows that DiCLIP achieves better segmentation performance even in challenging and fine-grained scenarios.
}}
  \label{fig.5}
  \vspace{-1.5em}
\end{figure*}

\begin{figure*}
  \centering
  \includegraphics[width=18.1cm]{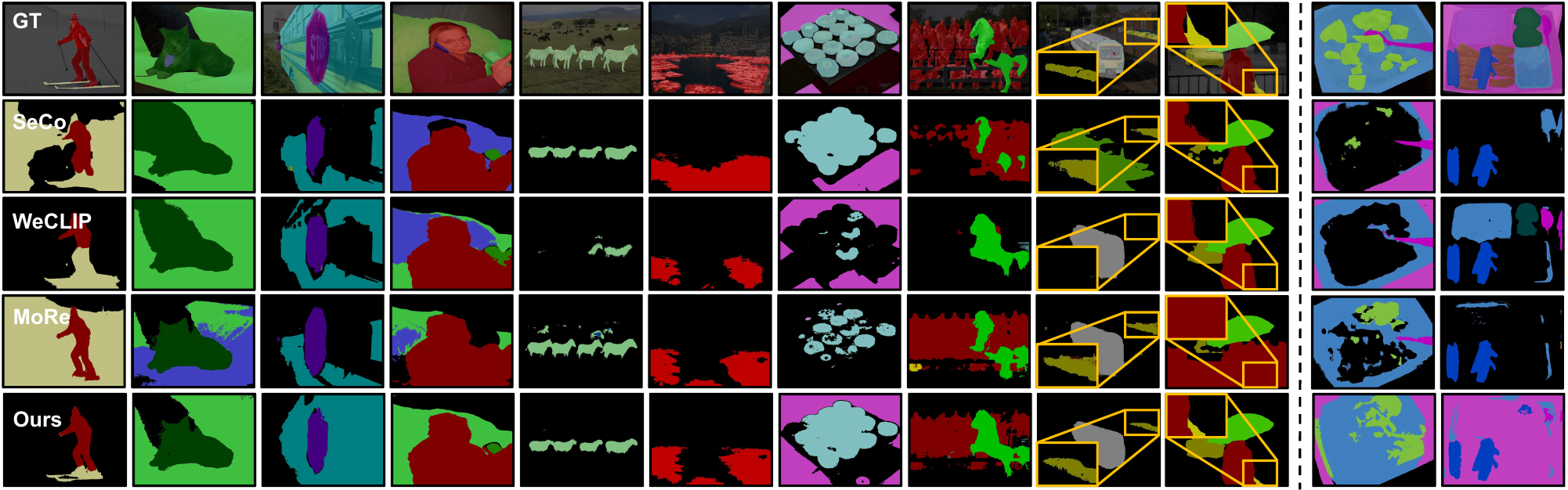}
  \caption{\hl{Qualitative segmentation comparisons on MS COCO 2014 with WeCLIP~{\cite{weakclip}}, SeCo~{\cite{seco}}, and MoRe~{\cite{MoreZW}}.
Small and off-center objects are highlighted with yellow rectangles, and failure cases are shown in the last two columns. Compared to other methods, DiCLIP produces more precise segmentation results.}}
  \label{fig.6}
  \vspace{-1.5em}
\end{figure*}

In addition, the qualitative comparisons of VOC and COCO are visualized in Fig.~\ref{fig.5} and Fig.~\ref{fig.6}, respectively. SeCo~\cite{seco}, WeCLIP~\cite{18}, and MoRe~\cite{MoreZW} are conducted as the competitors. \hl{By leveraging both CLIP and SD, our method produces more accurate predictions, even for small, off-center objects (highlighted by yellow rectangles). The last two columns present failure cases, highlighting the difficulty of extremely complex co-occurring regions. Nevertheless, DiCLIP still achieves better performance, as semantically pure prototypes in the cache reinforce category-consistent cues while suppressing irrelevant semantics, leading to more reliable predictions.}

\begin{table}[!t]
\centering
\caption{Segmentation comparisons on COCO. $Sup.$: supervision. $\mathcal{I}$: image-level labels. $\mathcal{SA}$: saliency maps. $\mathcal{L}$: Language.}
    \tablestyle{10.2pt}{1.1}
    \scalebox{1.}
    {
    \footnotesize
    \begin{tabular}{lccc}
    \toprule
    \multicolumn{1}{l|}{Method}                                       & \multicolumn{1}{c|}{Sup.}                               & \multicolumn{1}{c|}{Backbone}                                  & Val           \\ \midrule                               
    \multicolumn{1}{l|}{L2G~\cite{L2G} \tiny CVPR'2022}                                             & \multicolumn{1}{c|}{$\mathcal{I}+\mathcal{SA}$}                                  & \multicolumn{1}{c|}{RN101}                                                                    & 44.2          \\

    \multicolumn{1}{l|}{RCA~\cite{RCA} \tiny CVPR'2023}                                             & \multicolumn{1}{c|}{$\mathcal{I}+\mathcal{SA}$}                                  & \multicolumn{1}{c|}{RN38}                                                                     & 36.8          \\

    \multicolumn{1}{l|}{Mat-label\cite{matlabel} \tiny ICCV'2023}                                             & \multicolumn{1}{c|}{$\mathcal{I}+\mathcal{SA}$}                                  & \multicolumn{1}{c|}{RN101}                                                          & 45.6          \\
        \multicolumn{1}{l|}{\textcolor{black}{GSM\cite{GSM} \tiny AAAI'2021}}                                             & \multicolumn{1}{c|}{$\mathcal{I}$}                                  & \multicolumn{1}{c|}{VGG16}                                                                     & \textcolor{black}{28.4}          \\

    \multicolumn{1}{l|}{AFA~\cite{12} \tiny CVPR'2022}                                             & \multicolumn{1}{c|}{$\mathcal{I}$}                                    & \multicolumn{1}{c|}{MiT-B1}                                                                   & 38.9          \\

    \multicolumn{1}{l|}{FPR\cite{fpr} \tiny ICCV'2023}                                             & \multicolumn{1}{c|}{$\mathcal{I}$}                                  & \multicolumn{1}{c|}{RN101}                                                                     & 44.0          \\

    \multicolumn{1}{l|}{USAGE\cite{usage} \tiny ICCV'2023}                                             & \multicolumn{1}{c|}{$\mathcal{I}$}                                  & \multicolumn{1}{c|}{RN38}                                                                     & 44.3          \\
    
    \multicolumn{1}{l|}{OCR~\cite{OCR} \tiny CVPR'2023}                                             & \multicolumn{1}{c|}{$\mathcal{I}$}                                    & \multicolumn{1}{c|}{RN38}                                                                     & 42.5          \\

    \multicolumn{1}{l|}{BECO~\cite{BECO} \tiny CVPR'2023}                                            & \multicolumn{1}{c|}{$\mathcal{I}$}                                    & \multicolumn{1}{c|}{RN101}                                                                    & 45.1          \\
    
    \multicolumn{1}{l|}{DiG~\cite{DiG} \tiny ECCV'2024}                                             & \multicolumn{1}{c|}{$\mathcal{I}$}                                    & \multicolumn{1}{c|}{WRN38}                                                                    & 45.5          \\
    \multicolumn{1}{l|}{MCTformer+~\cite{33} \tiny TPAMI'2024}                                     & \multicolumn{1}{c|}{$\mathcal{I}$}                                    & \multicolumn{1}{c|}{RN38}                                                                     & 45.2          \\
    \multicolumn{1}{l|}{SSC~\cite{t1} \tiny TIP'2024}                                     & \multicolumn{1}{c|}{$\mathcal{I}$}                                    & \multicolumn{1}{c|}{RN101}                                                                     & 47.0          \\
    \multicolumn{1}{l|}{PLDA~\cite{PLDA} \tiny TIP'2024}                                     & \multicolumn{1}{c|}{$\mathcal{I}$}                                    & \multicolumn{1}{c|}{RN101}                                                                     & 44.7          \\
    \multicolumn{1}{l|}{DGRM~\cite{t1} \tiny TCSVT'2025}                                     & \multicolumn{1}{c|}{$\mathcal{I}$}                                    & \multicolumn{1}{c|}{RN101}                                                                     & 46.3          \\

    \multicolumn{1}{l|}{CLIMS~\cite{15} \tiny CVPR'2022}                                           & \multicolumn{1}{c|}{$\mathcal{I}+\mathcal{L}$}                                  & \multicolumn{1}{c|}{RN101}                                                                  & -             \\
    
    \multicolumn{1}{l|}{CLIP-ES~\cite{16} \tiny CVPR'2023}                                         & \multicolumn{1}{c|}{$\mathcal{I}+\mathcal{L}$}                                  & \multicolumn{1}{c|}{RN101}                                                                    & 45.4          \\
    
    \multicolumn{1}{l|}{PSDPM~\cite{PSDPM} \tiny CVPR'2024}                                           & \multicolumn{1}{c|}{$\mathcal{I}+\mathcal{L}$}                                  & \multicolumn{1}{c|}{RN101}                                                                  & 47.2          \\
    \multicolumn{1}{l|}{CPAL~\cite{37} \tiny CVPR'2024}                                            & \multicolumn{1}{c|}{$\mathcal{I}+\mathcal{L}$}                                  & \multicolumn{1}{c|}{RN101}                                                                    & 46.8          \\ 
    
    \multicolumn{1}{l|}{ToCo~\cite{22} \tiny CVPR'2023}                                            & \multicolumn{1}{c|}{$\mathcal{I}$}                                    & \multicolumn{1}{c|}{ViT-B}                                                                   & 42.3          \\
    \multicolumn{1}{l|}{DuPL~\cite{DuPL} \tiny CVPR'2024}                                            & \multicolumn{1}{c|}{$\mathcal{I}$}                                    & \multicolumn{1}{c|}{ViT-B}                                                                    & 44.6          \\
    \multicolumn{1}{l|}{SeCo~\cite{seco} \tiny CVPR'2024}                                            & \multicolumn{1}{c|}{$\mathcal{I}$}                                    & \multicolumn{1}{c|}{ViT-B}                                                                    & 46.7          \\
    \multicolumn{1}{l|}{\textcolor{black}{FFR~\cite{FFR} \tiny CVPR'2025}}                                            & \multicolumn{1}{c|}{$\mathcal{I}$}                                    & \multicolumn{1}{c|}{ViT-B}                                                                    & \textcolor{black}{46.8}          \\
    \multicolumn{1}{l|}{\textcolor{black}{PCRE~\cite{PCRE} \tiny CVPR'2025}}                                            & \multicolumn{1}{c|}{$\mathcal{I}$}                                    & \multicolumn{1}{c|}{ViT-B}                                                                    & \textcolor{black}{47.2}          \\
    \multicolumn{1}{l|}{\textcolor{black}{MuP-VSS~\cite{MuP-VSS} \tiny CVPR'2025}}                                            & \multicolumn{1}{c|}{$\mathcal{I}$}                                    & \multicolumn{1}{c|}{ViT-B}                                                                    & \textcolor{black}{46.6}          \\
    \multicolumn{1}{l|}{\textcolor{black}{ToMa~\cite{ToMa} \tiny ICCV'2025}}                                            & \multicolumn{1}{c|}{$\mathcal{I}$}                                    & \multicolumn{1}{c|}{ViT-B}                                                                    & \textcolor{black}{43.2}          \\
    \multicolumn{1}{l|}{\textcolor{black}{MoRe~\cite{MoreZW} \tiny AAAI'2025}}                                            & \multicolumn{1}{c|}{$\mathcal{I}$}                                    & \multicolumn{1}{c|}{ViT-B}                                                                    & \textcolor{black}{47.4}          \\
    \multicolumn{1}{l|}{DIAL~\cite{DIAL} \tiny ECCV'2024}                                            & \multicolumn{1}{c|}{$\mathcal{I}+\mathcal{L}$}                                  & \multicolumn{1}{c|}{ViT-B}                                                            & 44.4          \\
    \multicolumn{1}{l|}{WeakCLIP~\cite{weakclip} \tiny IJCV'2024}                                            & \multicolumn{1}{c|}{$\mathcal{I}+\mathcal{L}$}                                  & \multicolumn{1}{c|}{MiT-B}                                                            & 47.4          \\
    \multicolumn{1}{l|}{WeCLIP~\cite{18} \tiny CVPR'2024}                                          & \multicolumn{1}{c|}{$\mathcal{I}+\mathcal{L}$}                                  & \multicolumn{1}{c|}{ViT-B}                                                                  & 47.1          \\
    \multicolumn{1}{l|}{\textcolor{black}{POT~\cite{POT} \tiny CVPR'2025}}                                          & \multicolumn{1}{c|}{$\mathcal{I}+\mathcal{L}$}                                  & \multicolumn{1}{c|}{ViT-B}                                                                  & \textcolor{black}{47.9}          \\
    \rowcolor[HTML]{EFEFEF} 
    \multicolumn{1}{l|}{\cellcolor[HTML]{EFEFEF}\textbf{DiCLIP(w/o CRF)}} & \multicolumn{1}{c|}{\cellcolor[HTML]{EFEFEF}\textbf{$\mathcal{I}+\mathcal{L}$}} & \multicolumn{1}{c|}{\cellcolor[HTML]{EFEFEF}\textbf{ViT-B}}  & \textbf{47.6} \\
    \rowcolor[HTML]{EFEFEF} 
    \multicolumn{1}{l|}{\cellcolor[HTML]{EFEFEF}\textbf{DiCLIP (Ours)}}          & \multicolumn{1}{c|}{\cellcolor[HTML]{EFEFEF}\textbf{$\mathcal{I}+\mathcal{L}$}} & \multicolumn{1}{c|}{\cellcolor[HTML]{EFEFEF}\textbf{ViT-B}}  & \textbf{48.7} \\ \bottomrule
    \end{tabular}
    }
    \label{tab.2}
   \vspace{-1.em}
\end{table}

\begin{table}[t!]
\centering
\caption{CAM seed without post-processing on VOC train set. $\mathcal{M}$: multi-staged.$\dagger$: our reproduction following official code.}
    \vspace{-0.2em}
    \tablestyle{6.8pt}{1.12}
    \scalebox{1.}
    {
    \footnotesize
    \begin{tabular}{@{}lcccc}
    \toprule
    \multicolumn{1}{l|}{Method}                                       & \multicolumn{1}{c|}{Type}                               & \multicolumn{1}{c|}{Sup.}                                  & Backbone    & Train       \\ \midrule 
    \multicolumn{5}{l}{\textit{\textbf{Training-free WSSS methods.}}}                                                                                                                                                                                                              \\
    \multicolumn{1}{l|}{Diffsegmenter~\cite{diffsegmenter} \tiny TIP'2025}                                 & \multicolumn{1}{c|}{$\mathcal{S}$}                                  & \multicolumn{1}{c|}{$\mathcal{I}$}                                  & \multicolumn{1}{c|}{SD}                                  & 70.5          \\
    \multicolumn{1}{l|}{CLIP-ES~\cite{16} \tiny CVPR'2023}                                 & \multicolumn{1}{c|}{$\mathcal{S}$}                                  & \multicolumn{1}{c|}{$\mathcal{I}+\mathcal{L}$}                                  & \multicolumn{1}{c|}{ViT-B}                                  & 70.8          \\
    \rowcolor[HTML]{EFEFEF} 
    \multicolumn{1}{l|}{\cellcolor[HTML]{EFEFEF}\textbf{DiCLIP ($M_{t}$})} & \multicolumn{1}{c|}{\cellcolor[HTML]{EFEFEF}\textbf{$\mathcal{S}$}} & \multicolumn{1}{c|}{\cellcolor[HTML]{EFEFEF}\textbf{$\mathcal{I}+\mathcal{L}$}} & \multicolumn{1}{c|}{\cellcolor[HTML]{EFEFEF}\textbf{ViT-B}} & \textbf{72.0} \\
    \rowcolor[HTML]{EFEFEF} 
    \multicolumn{1}{l|}{\cellcolor[HTML]{EFEFEF}\textbf{DiCLIP ($M_{es}$})} & \multicolumn{1}{c|}{\cellcolor[HTML]{EFEFEF}\textbf{$\mathcal{S}$}} & \multicolumn{1}{c|}{\cellcolor[HTML]{EFEFEF}\textbf{$\mathcal{I}+\mathcal{L}$}} & \multicolumn{1}{c|}{\cellcolor[HTML]{EFEFEF}\textbf{ViT-B}} & \textbf{74.0} \\ \midrule
    \multicolumn{5}{l}{\textit{\textbf{Training-required WSSS methods.}}}        \\
    \multicolumn{1}{l|}{ReCAM~\cite{30} \tiny CVPR'2022}                                   & \multicolumn{1}{c|}{$\mathcal{M}$}                                  & \multicolumn{1}{c|}{$\mathcal{I}$}                                    & \multicolumn{1}{c|}{RN101}                                  & 54.8          \\
    \multicolumn{1}{l|}{FPR~\cite{fpr} \tiny CVPR'2023}                                     & \multicolumn{1}{c|}{$\mathcal{M}$}                                  & \multicolumn{1}{c|}{$\mathcal{I}$}                                    & \multicolumn{1}{c|}{RN101}                                  & 63.8          \\
    \multicolumn{1}{l|}{PLDA~\cite{PLDA} \tiny TIP'2024}                              & \multicolumn{1}{c|}{$\mathcal{M}$}                                  & \multicolumn{1}{c|}{$\mathcal{I}$}                                    & \multicolumn{1}{c|}{RN101}                                   & 62.5          \\
    \multicolumn{1}{l|}{MCTformer+~\cite{33} \tiny TPAMI'2024}                              & \multicolumn{1}{c|}{$\mathcal{M}$}                                  & \multicolumn{1}{c|}{$\mathcal{I}$}                                    & \multicolumn{1}{c|}{RN38}                                   & 68.8          \\
    \multicolumn{1}{l|}{SFC~\cite{34} \tiny AAAI'2024}                                     & \multicolumn{1}{c|}{$\mathcal{M}$}                                  & \multicolumn{1}{c|}{$\mathcal{I}$}                                    & \multicolumn{1}{c|}{RN101}                                  & 64.7          \\
    \multicolumn{1}{l|}{CTI~\cite{35} \tiny CVPR'2024}                                     & \multicolumn{1}{c|}{$\mathcal{M}$}                                  & \multicolumn{1}{c|}{$\mathcal{I}$}                                    & \multicolumn{1}{c|}{RN101}                                  & 69.5          \\
    \multicolumn{1}{l|}{AFA~\cite{12} \tiny CVPR'2022}                                     & \multicolumn{1}{c|}{$\mathcal{S}$}                                  & \multicolumn{1}{c|}{$\mathcal{I}$}                                    & \multicolumn{1}{c|}{MiT-B1}                                 & 65.0          \\
    \multicolumn{1}{l|}{ViT-PCM~\cite{ViT-PCM} \tiny ECCV'2022}                                 & \multicolumn{1}{c|}{$\mathcal{S}$}                                  & \multicolumn{1}{c|}{$\mathcal{I}$}                                    & \multicolumn{1}{c|}{ViT-B}                                  & 67.7          \\
    \multicolumn{1}{l|}{$\dagger$ToCo~\cite{22} \tiny CVPR'2023}                                    & \multicolumn{1}{c|}{$\mathcal{S}$}                                  & \multicolumn{1}{c|}{$\mathcal{I}$}                                    & \multicolumn{1}{c|}{ViT-B}                                  & 71.6          \\
    \multicolumn{1}{l|}{$\dagger$DuPL~\cite{DuPL} \tiny CVPR'2024}                                    & \multicolumn{1}{c|}{$\mathcal{S}$}                                  & \multicolumn{1}{c|}{$\mathcal{I}$}                                    & \multicolumn{1}{c|}{ViT-B}                                  & 75.0          \\
    \multicolumn{1}{l|}{SeCo~\cite{seco} \tiny CVPR'2024}                                    & \multicolumn{1}{c|}{$\mathcal{S}$}                                  & \multicolumn{1}{c|}{$\mathcal{I}$}                                    & \multicolumn{1}{c|}{ViT-B}                                  & 74.8          \\
    \multicolumn{1}{l|}{\textcolor{black}{FFR~\cite{FFR} \tiny CVPR'2025}}                                    & \multicolumn{1}{c|}{$\mathcal{S}$}                                  & \multicolumn{1}{c|}{$\mathcal{I}$}                                    & \multicolumn{1}{c|}{ViT-B}                                  &\textcolor{black}{-}          \\
    \multicolumn{1}{l|}{\textcolor{black}{PCRE~\cite{PCRE} \tiny CVPR'2025}}                                    & \multicolumn{1}{c|}{$\mathcal{S}$}                                  & \multicolumn{1}{c|}{$\mathcal{I}$}                                    & \multicolumn{1}{c|}{ViT-B}                                  &\textcolor{black}{-}          \\
    \multicolumn{1}{l|}{\textcolor{black}{ToMa~\cite{ToMa} \tiny ICCV'2025}}                                    & \multicolumn{1}{c|}{$\mathcal{S}$}                                  & \multicolumn{1}{c|}{$\mathcal{I}$}                                    & \multicolumn{1}{c|}{ViT-B}                                  &\textcolor{black}{-}          \\
    \multicolumn{1}{l|}{\textcolor{black}{MoRe~\cite{MoreZW} \tiny AAAI'2025}}                                    & \multicolumn{1}{c|}{$\mathcal{S}$}                                  & \multicolumn{1}{c|}{$\mathcal{I}$}                                    & \multicolumn{1}{c|}{ViT-B}                                  &\textcolor{black}{77.0}          \\
    \multicolumn{1}{l|}{\textcolor{black}{MuP-VSS~\cite{MuP-VSS} \tiny CVPR'2025}}                                    & \multicolumn{1}{c|}{$\mathcal{M}$}                                  & \multicolumn{1}{c|}{$\mathcal{I}$}                                    & \multicolumn{1}{c|}{ViT-B}                                  &\textcolor{black}{71.7}          \\
    \multicolumn{1}{l|}{DiG~\cite{DiG} \tiny ECCV'2024}                                    & \multicolumn{1}{c|}{$\mathcal{M}$}                                  & \multicolumn{1}{c|}{$\mathcal{I}$}                                    & \multicolumn{1}{c|}{DiT-S}                                  & 69.3          \\
    \multicolumn{1}{l|}{CLIMS~\cite{15} \tiny CVPR'2022}                                   & \multicolumn{1}{c|}{$\mathcal{M}$}                                  & \multicolumn{1}{c|}{$\mathcal{I}+\mathcal{L}$}                                  & \multicolumn{1}{c|}{RN101}                                  & 56.6          \\
    \multicolumn{1}{l|}{POLE~\cite{promptclass} \tiny WACV'2023}                                    & \multicolumn{1}{c|}{$\mathcal{M}$}                                  & \multicolumn{1}{c|}{$\mathcal{I}+\mathcal{L}$}                                  & \multicolumn{1}{c|}{RN50}                                   & 59.0          \\
    \multicolumn{1}{l|}{CPAL~\cite{37} \tiny CVPR'2024}                                    & \multicolumn{1}{c|}{$\mathcal{M}$}                                  & \multicolumn{1}{c|}{$\mathcal{I}+\mathcal{L}$}                                  & \multicolumn{1}{c|}{ViT-B}                                  & 71.9          \\
    \multicolumn{1}{l|}{$\dagger$WeakCLIP~\cite{weakclip} \tiny IJCV'2024}                                    & \multicolumn{1}{c|}{$\mathcal{M}$}                                  & \multicolumn{1}{c|}{$\mathcal{I}+\mathcal{L}$}                                  & \multicolumn{1}{c|}{ViT-B}                                  & 73.7          \\
    \multicolumn{1}{l|}{DIAL~\cite{DIAL} \tiny ECCV'2024}                                    & \multicolumn{1}{c|}{$\mathcal{S}$}                                  & \multicolumn{1}{c|}{$\mathcal{I}+\mathcal{L}$}                                  & \multicolumn{1}{c|}{ViT-B}                                  & 75.2          \\
    \multicolumn{1}{l|}{$\dagger$WeCLIP~\cite{18} \tiny CVPR'2024}                                  & \multicolumn{1}{c|}{$\mathcal{S}$}                                  & \multicolumn{1}{c|}{$\mathcal{I}+\mathcal{L}$}                                  & \multicolumn{1}{c|}{ViT-B}                                  & 75.4          \\
    \multicolumn{1}{l|}{\textcolor{black}{POT~\cite{POT} \tiny CVPR'2025}}                                  & \multicolumn{1}{c|}{$\mathcal{S}$}                                  & \multicolumn{1}{c|}{$\mathcal{I}+\mathcal{L}$}                                  & \multicolumn{1}{c|}{ViT-B}                                  & \textcolor{black}{75.0}          \\
    \rowcolor[HTML]{EFEFEF} 
    \multicolumn{1}{l|}{\cellcolor[HTML]{EFEFEF}\textbf{DiCLIP (Ours-$M_{ed}$)}}  & \multicolumn{1}{c|}{\cellcolor[HTML]{EFEFEF}\textbf{$\mathcal{S}$}} & \multicolumn{1}{c|}{\cellcolor[HTML]{EFEFEF}\textbf{$\mathcal{I}+\mathcal{L}$}} & \multicolumn{1}{c|}{\cellcolor[HTML]{EFEFEF}\textbf{ViT-B}} & \textbf{78.2} \\ \bottomrule
    \end{tabular}
    }
   \label{tab.3}
   \vspace{-1.5em}
\end{table}

\begin{figure*}
  \centering
  \includegraphics[width=18.1cm]{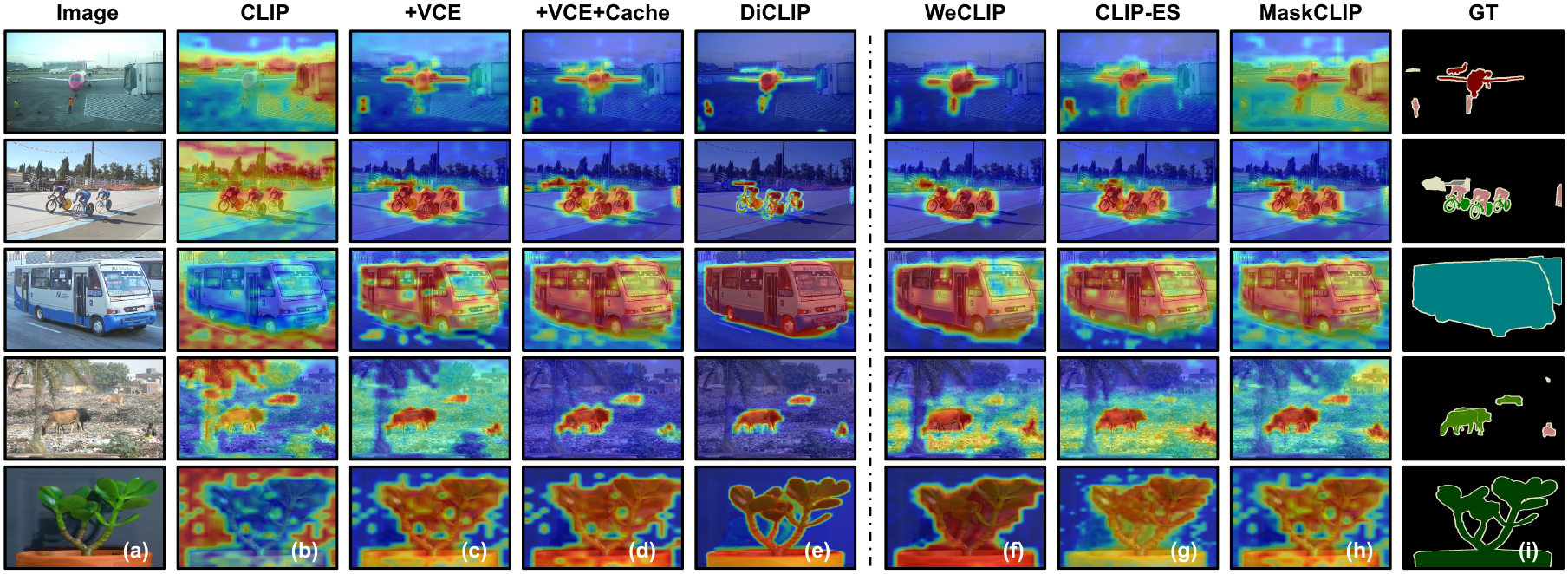}
   \caption{\hl{CAM visualizations on VOC train set, validating the efficacy of our module.} (a) Image. (b-e) Qualitative ablation study of our key components. (e-h) Comparisons between (e) DiCLIP and recent CLIP-based counterparts, i.e., (f) WeCLIP~\cite{18}, (g) CLIP-ES~\cite{16}, and MaskCLIP~\cite{maskclip}. (i) Ground truth mask.}
   \label{fig.7}
  \vspace{-.5em}
\end{figure*}

{\bf Evaluation of CAM Seeds.}~Table~\ref{tab.3} reports the initial CAM seeds on VOC train set. Based on CLIP and SD, our method generates CAMs in both training-free and training-efficient manners. By only enhancing CLIP's image encoder with SD, DiCLIP generates patch-text CAM $M_{t}$ with $72.0 \%$ mIoU. With the static knowledge retrieval from cache model, the performance increases to $74.0 \%$, which surpasses CLIP-based~\cite{16}, SD-based~\cite{diffsegmenter}, and even most of training-required methods. When the adapter is initiated with our cache model, we further make the CAM generation learnable and achieve $78.2 \%$ mIoU. It significantly outperforms recent CLIP-based SOTAs at least by $2.8 \%$ mIoU. Moreover, the qualitative comparisons are drawn in Fig.~\ref{fig.7} (e-h). Our method generates better localization maps than recent CLIP-based methods, showing the efficacy of enhancing CLIP with SD.
\subsection{Experiment Settings}
{\bf Datasets and Evaluation Metrics.} DiCLIP is evaluated on PASCAL VOC 2012~\cite{VOC} and MS COCO 2014~\cite{COCO} following prior practice~\cite{weakclip,18,seco}. PASCAL VOC has $20$ foreground and $1$ background classes, where $10,582$, $1,449$, and $1,456$ images are augmented for train, val, and test set. For MS COCO dataset, it consists of $80$ foreground and $1$ background classes, with $82,081$ images in train set and $40,137$ images in val set. 
For evaluation, Mean Intersection-Over-Union (mIoU) serves as the primary metric. Additionally, the confusion ratio (CR)~\cite{seco} is used to assess segmentation precision, calculated as false positives divided by true positives.

{\bf Implementation Details.} 
DiCLIP adopts ViT-based~\cite{ViT} CLIP and Stable Diffusion 2.1~\cite{sd}, both kept frozen during training. In VCE module, we extract attention maps from self-attention layers of UNet decoder and average them among the head dimension following~\cite{diffsegmenter}. {To preserve SD’s diversity while ensuring scale consistency with CLIP, we interpolate SD’s queries and keys to match CLIP’s feature size, generating the SD attention maps.} $B=9$ groups of attention maps are clustered in Eq.~\ref{eq:4} and $R=3$ times of ACR refinement are conducted in Eq.~\ref{eq:6}. $\alpha$ is set to $1$ in Eq.~\ref{eq:7} by default. We conduct VCE in the last $L=3$ attention layers of CLIP. 

{In TSA, we generate $50$ single-class images (resolution: $384 \times 384$) per category using $45$ inference steps with a guidance scale of $10$. The prompt template is: \textit{'a realistic photograph of a fully visible, entire \{category\} with natural colors, centered in the image, with a clear and distinct background, high color contrast, and the \{category\} not occupying the entire frame, ensuring it is completely visible without cropping.'}, where \{category\} denotes the queried class. We avoid complex background prompts, as overly intricate backgrounds may degrade the purity of the foreground cache. Deduplication strategy is also not adopted, as our clustering operation naturally removes duplicates.} The number of clustered features $E$ is set to $10$ and $5$ for VOC and COCO. With $N$ learnable key-value prompts, we build a total $312$ key-value pairs for VOC and $512$ pairs for COCO. Following~\cite{18}, we use a Transformer-based decoder head. Frozen features from each layer of raw CLIP are extracted for segmentation. The loss weight $\gamma$ is set as $0.1$. During training, AdamW optimizer is used following~\cite{seco, DuPL, 22}. The learning rate is $2e-4$ with a weight decay of $1e-2$. The iteration is set as $20,000$ for VOC and $100,000$ for COCO. All experiments are implemented with NVIDIA RTX 3090.
\begin{table}[!t]
\centering
\vspace{-1.0em}
    \caption{Ablative study of our key components on VOC val set for segmentation. P.C.: Positive Cache. N.C.: Negative Cache. }
   \vspace{-.5em}
    \tablestyle{4.2pt}{1.}
    \scalebox{1.}
    {
   \begin{tabular}{l|ccccccc}
    \toprule
    Conditions                & VCE & P.C. & N.C. & Adapter & Precision & Recall & mIoU \\ \midrule
    Baseline              &     &      &     &      &18.8           &21.3        & 12.1 \\ \midrule
    \multirow{4}{*}{Ours} &\pmb{$\checkmark$}     &      &     &      & 84.3          &82.9        & 72.2 \\
                          &\pmb{$\checkmark$}     &\pmb{$\checkmark$}      &     &      & 83.8      & 85.4   & 73.6 \\
                          &\pmb{$\checkmark$}     &\pmb{$\checkmark$}      &\pmb{$\checkmark$}     &      & 84.8      & 85.4   & 74.4 \\
    \rowcolor[HTML]{EFEFEF} 
                          
                          &\pmb{$\checkmark$}     &\pmb{$\checkmark$}      &\pmb{$\checkmark$}     &\pmb{$\checkmark$}      & \textbf{85.2}      & \textbf{87.8}   & \textbf{77.1} \\ \bottomrule
    \end{tabular}
    }
    \label{tab.4}
   \vspace{-0.8em}
\end{table}

\begin{table}[!t]
\centering
    \caption{Ablation of key-value cache model on VOC val set.}
   \vspace{-.5em}
    \tablestyle{2.8pt}{1.1}
    \scalebox{1.}
    {
    \begin{tabular}{l|cccccc}
    \toprule
    Conditions                & Clustering & Text Values & Initiation & Precision & Recall & mIoU  \\ \midrule
    \multirow{4}{*}{Ours} & \pmb{$\checkmark$}                & \pmb{$\checkmark$}                 &         & 83.3  & 88.5  & 76.0 \\
                          & \pmb{$\checkmark$}                &                 &\pmb{$\checkmark$}          & 84.5  & 87.3  & 76.2 \\
                          &                 &\pmb{$\checkmark$}                 &\pmb{$\checkmark$}          & 83.4 & 87.8 & 75.3\\
    \rowcolor[HTML]{EFEFEF} 
                          
                          & \pmb{$\checkmark$}                & \pmb{$\checkmark$}                & \pmb{$\checkmark$}        & \textbf{85.2} & \textbf{87.8} & \textbf{77.1}\\ \bottomrule
    \end{tabular}
    }
    \label{tab.5}
   \vspace{-1.5em}
\end{table}

\subsection{Ablation Studies}
{\bf Effectiveness of Key Components.} The quantitative ablation studies of our method are reported in Table \ref{tab.4}. CRF is not used. The baseline refers to using the raw ViT-based CLIP to directly generate CAM following Eq.~\ref{eq:1} and train the decoder end-to-end. Without our modules, it only achieves $12.1\%$ mIoU on PASCAL VOC val set. \hl{Incorporating the proposed VCE module to diversify CLIP’s overly smooth attention leads to a significant performance gain, improving mIoU to $72.2\%$. This validates our first assumption (\emph{Spatial Consistency}), demonstrating that enriching CLIP with spatially coherent attention effectively benefits dense prediction. Furthermore, with the proposed Positive Cache (P.C.), DiCLIP injects dense semantic knowledge into patch-text alignment and achieves $73.6\%$ mIoU, supporting our second assumption (\emph{Generative Validity}). When the Negative Cache (N.C.) is additionally employed, the performance further improves by $0.8\%$ mIoU, highlighting the benefit of contrastive, patch-wise knowledge retrieval from semantically pure prototypes.} In addition, when the adapter is initiated with our key-value cache model, DiCLIP enables the CAM generation to a dynamic process and finally achieves $77.1\%$ mIoU. \hl{These ablation studies validate our assumptions and prove that our modules effectively enhance CLIP’s dense capability for WSSS.}

To convincingly illustrate the efficacy of our modules, we further conduct ablative visualizations in Fig.~\ref{fig.7} (b-e). Fig.~\ref{fig.7} (b) shows that the CLIP baseline cannot generate reasonable localization maps due to its limited dense knowledge across vision and text modalities. Our VCE module alleviates this issue by diversifying visual features and enriching fine-grained details with knowledge from SD, as shown in Fig.~\ref{fig.7} (c). In addition, we also incorporate a static key-value cache model to enhance CLIP's patch-text representations, which generates more complete CAM, as illustrated in Fig.~\ref{fig.7} (d). To further improve the completeness and precision of CAM, a dynamic adapter is further built from the cache model and enables learnable CAM generation, as seen in Fig.~\ref{fig.7} (e). Both quantitative and qualitative ablation experiments provide strong evidence to validate the effectiveness of our modules.

{\bf Efficacy of Key-value Cache Model.} Table~\ref{tab.5} specifically analyzes different strategies when we maintain the key-value cache model. 'Initiation' means that we use the key-value pairs to initiate the weights of the adapter. When we build the adapter with random initiation, the performance drops $1.1 \%$ mIoU. One potential reason is that our cache model contains rich priors for category representations. Such prior efficiently helps the adapter learn reliable class knowledge for CAM generation. It is also noted that even with the randomly-initiated adapter, DiCLIP gains improvement compared to the situation where no adapter is built ($74.4 \%$ mIoU in Table~\ref{tab.4}). 'Text Values' denotes that we use text embeddings to assign different weights to Values instead of directly taking the one-hot labels as Values. Incorporating it improves the performance from $76.2 \%$ to $77.1 \%$, as it works by enabling cache model to capture intra- and inter-class differences of Keys. 'Clustering' means that we store the clustered image features as Keys instead of caching all generated image features. It is reported that the performance drops to $75.3 \%$ when we exclude it. It could be explained that such an operation selects the most representative features as Keys. It effectively reduces the irrelevant noise and duplicated samples during knowledge retrieval. The ablative experiments clearly validate the advantage of our cache model to enrich dense knowledge of CLIP.
\begin{figure}[!t]
  \centering
  \includegraphics[width=\linewidth]{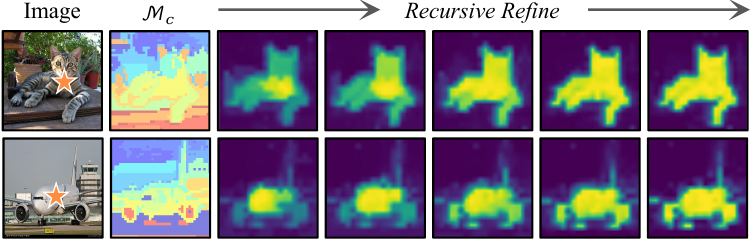}
   \caption{\hl{Evolution of attention maps throughout recursive refinement on VOC train set, illustrating progressively improved SD attention maps.}}
   \label{fig.8}
   \vspace{-1.0em}
\end{figure}
\begin{table*}[]
  \centering
  \caption{{Ablative analysis of hyper-parameters in our key components on VOC val set. M: quality of CAM seeds. Seg.: semantic predictions. Post-processing CRF is not used. Attn.: Attention. num.: Number.}}
  \vspace{-1.em}
  \subfloat[{ATTN GROUPS $B$.}]
  {
    \centering
    \begin{minipage}[b]{0.24\linewidth}{
    \begin{center}
    \tablestyle{9pt}{1.05}
    \scalebox{0.8}{
    \begin{tabular}{l|cc}
    \toprule
    \multicolumn{1}{c|}{Attn. groups $B$} & M   & Seg. \\ \midrule
    3          & 75.7     & 74.9     \\
    6          & 77.4     & 76.4      \\
    \rowcolor[HTML]{EFEFEF} 
    \textbf{9} & \textbf{78.2} & \textbf{77.1} \\
    12          & {77.7}      & 76.7     \\ \bottomrule
    \end{tabular}
    }
    \end{center}
    }
    \end{minipage}
   }
\subfloat[{ATTN LAYERS $L$.}]
  {
    \centering
    \begin{minipage}[b]{0.24\linewidth}{
    \begin{center}
    \tablestyle{9pt}{1.05}
    \scalebox{0.8}{
    \begin{tabular}{l|cc}
    \toprule
    Attn. layers $L$                 & M    & Seg. \\ \midrule
    1                        &71.7  & 71.4 \\
    \rowcolor[HTML]{EFEFEF} 
    {\color[HTML]{000000}\textbf{3}} & {\color[HTML]{000000}\textbf{78.2}}  & {\color[HTML]{000000}\textbf{77.1}} \\
    4                        &77.4  & 76.9 \\
    6                        &76.9  & 75.4 \\ \bottomrule
    \end{tabular}
    }
    \end{center}
    }
    \end{minipage}
   }
   \subfloat[{ATTN THRESHOLD $\epsilon$.}]
  {
    \centering
    \begin{minipage}[b]{0.24\linewidth}{
    \begin{center}
    \tablestyle{10pt}{1.05}
    \scalebox{0.8}{
        \begin{tabular}{l|cc}
    \toprule
    Attn Thre  $\epsilon$ & M             & Seg.           \\ \midrule
    1e-5               & 77.8          & 76.7          \\ 
    1e-4               & 78.0          & 77.0          \\
    \rowcolor[HTML]{EFEFEF} 
    \textbf{5e-4}      & \textbf{78.2} & \textbf{77.1} \\
    1e-3               & 77.7          & 76.6          \\ \bottomrule
    \end{tabular}
    }
    \end{center}
    }
    \end{minipage}
  }
     \subfloat[{ATTN WEIGHT $\alpha$.}]
  {
    \centering
    \begin{minipage}[b]{0.24\linewidth}{
    \begin{center}
    \tablestyle{8pt}{1.05}
    \scalebox{0.8}{
    \begin{tabular}{l|cc}
    \toprule
    Attn weight $\alpha$  & M             & Seg.           \\ \midrule
    0.1          & 76.1          & 75.3          \\
    0.5          & 77.6          & 76.8          \\
    \rowcolor[HTML]{EFEFEF} 
    \textbf{1.0} & \textbf{78.2} & \textbf{77.1} \\
    1.5          & 77.8          & 76.9          \\ \bottomrule
    \end{tabular}
    }
    \end{center}
    }
    \end{minipage}
    }\\
\subfloat[{CLASS CENTROID $E$.}]
  {
    \centering
    \begin{minipage}[b]{0.24\linewidth}{
    \begin{center}
    \tablestyle{8pt}{1.05}
    \scalebox{0.8}{

    \begin{tabular}{l|cc}
    \toprule
    Class centroid $E$    & M    & Seg. \\ \midrule
    6                        &77.3  & 76.2  \\
    8                       &77.4  &76.6  \\
    \rowcolor[HTML]{EFEFEF} 
    {\color[HTML]{000000}\textbf{10}} & {\color[HTML]{000000}\textbf{78.2}}  & {\color[HTML]{000000}\textbf{77.1}} \\
    12                        &77.9  &76.8   \\ \bottomrule
    \end{tabular}
    }
    \end{center}
    }
    \end{minipage}
  }
  \subfloat[{PROMPT NUMBER $N$.}]
  {
    \centering
    \begin{minipage}[b]{0.24\linewidth}{
    \begin{center}
    \tablestyle{8pt}{1.05}
    \scalebox{0.8}{
    \begin{tabular}{l|cc}
    \toprule
    Prompt num. $N$                  & M    & Seg. \\ \midrule
    0 (220)                        &77.6  &76.6  \\
    36 (256)                        & 77.9 & 76.9 \\
    \rowcolor[HTML]{EFEFEF} 
    {\color[HTML]{000000}\textbf{92 (312)}} & {\color[HTML]{000000}\textbf{78.2}}  & {\color[HTML]{000000}\textbf{77.1}} \\
    292 (512)                        &77.3  &76.4 \\ \bottomrule
    \end{tabular}
    }
    \end{center}
    }
    \end{minipage}
  }
  \subfloat[{CACHE WEIGHT $\beta$.}]
  {
    \centering
    \begin{minipage}[b]{0.24\linewidth}{
    \begin{center}
    \tablestyle{8pt}{1.05}
    \scalebox{0.8}{
    \begin{tabular}{l|cc}
    \toprule
    Cache weight $\beta$  & M    & Seg. \\ \midrule
    0.1                       & 75.9 & 75.1 \\
    0.3                        &76.9  & 76.4 \\
    \rowcolor[HTML]{EFEFEF} 
    {\color[HTML]{000000}\textbf{0.5}} & {\color[HTML]{000000}\textbf{78.2}}  & {\color[HTML]{000000}\textbf{77.1}} \\ 
    1.0                        &77.4  & 76.8 \\
 \bottomrule
    \end{tabular}
    }
    \end{center}
    }
    \end{minipage}
  }
  \subfloat[{LOSS WEIGHT $\gamma$.}]
  {
    \centering
    \begin{minipage}[b]{0.24\linewidth}{
    \begin{center}
    \tablestyle{8pt}{1.05}
    \scalebox{0.8}{
    \begin{tabular}{l|cc}
    \toprule
    \multicolumn{1}{c|}{Loss weight $\gamma$}  & M    & Seg. \\ \midrule
    0.05                        &77.0  & 76.6 \\
    \rowcolor[HTML]{EFEFEF} 
    {\color[HTML]{000000}\textbf{0.1}} & {\color[HTML]{000000}\textbf{78.2}}  & {\color[HTML]{000000}\textbf{77.1}} \\ 
    0.5                              & 76.8 & 76.0 \\
    1.0                              &76.1  &75.3  \\\bottomrule
    \end{tabular}
    }
    \end{center}
    }
    \end{minipage}
  }
  \vspace{-1.5 em}
\label{tab.6}
\end{table*}%

\begin{table}[]
\centering
\caption{Ablation study of ACR refinement number $R$ on VOC val set.}
   \vspace{-.5em}
    \tablestyle{6.5pt}{1}
    \scalebox{1.}
    {
    \footnotesize
    \begin{tabular}{l|cccccc}
    \toprule
    Number of Iter. & w/o ACR & 1 & 2 & \textbf{3}  & 4 & 5 \\ \midrule
    mIoU           & 75.1   &75.8    &76.2    & \textbf{77.1} & 77.0    & 76.5    \\ \bottomrule
    \end{tabular}
    }
   \label{tab.7}
   \vspace{-1em}
\end{table}
{\bf Efficacy of Attention Clustering Refinement.} Since the frozen attention maps from SD may lack clear relations in subtle regions, we introduce ACR module to iteratively refine them. It enhances correlations among the same semantics while suppressing irrelevant ones, generating better attention maps. In Table~\ref{tab.7}, we evaluate its efficacy by varying the iteration number. Without ACR, directly adding attention maps to CLIP reduces performance from $77.1\%$ to $75.1\%$ mIoU. Excessive iterations lead to a decline as well, likely because the refined correlations in attention maps become overly rigid. It reports that DiCLIP achieves the best performance ($77.1\%$ mIoU) at $3$ iterations. {In addition, Fig.{~\ref{fig.8}} illustrates the evolution of attention maps throughout the recursive refinement process. The results validate the efficacy of our recursive ACR and its superiority over plain attention maps from SD.}

{\bf {Ablative Analysis of Hyper-parameters}.} In Table~\ref{tab.6}, we perform the ablation study to analyze the impact of hyper-parameters in the proposed modules. {Grid search strategy is used based on the segmentation and CAM performance on VOC. \hl{In Table{~\ref{tab.6}} (a-d), we assess the robustness of our VCE module by evaluating its performance sensitivity to key hyper-parameters}, including the number of attention clustering groups $B$, the number of enhanced attention layers $L$ in CLIP, the attention threshold $\epsilon$ for filtering noisy values, and the attention weight $\alpha$ in Eq.~\ref{eq:7}. The group number $B$ determines the granularity of clustered semantics. As shown in Table~{\ref{tab.6}} (a), the performance drops significantly when $B=3$, likely because many irrelevant tokens are clustered together, introducing substantial noise in the affinity relations. Our results indicate that DiCLIP achieves the best performance when $B=9$. In Table~{\ref{tab.6} (b), the performance of DiCLIP improves significantly as we increase $L$ from 1 to 3. This is because applying VCE from the intermediate layers of CLIP progressively enhances feature diversity, which in turn benefits CAM generation. In Table~{\ref{tab.6}} (c), DiCLIP maintains stable performance across different values of $\epsilon$, demonstrating the robustness of the proposed ACR module. In Table~{\ref{tab.6}} (d), a significant performance drop is observed when $\alpha=0.1$, as a small $\alpha$ weakens the calibration effect from SD's attention. As $\alpha$ increases, the performance gradually improves and reaches the best result at $\alpha=1.0$. \hl{Fig.{~\ref{fig.9}} further visualizes the impact of varying $\alpha$ on the attention distribution, providing more intuitive evidence of the effectiveness of our VCE module.}
\begin{figure}[!t]
  \centering
   \vspace{.1em}
  \includegraphics[width=8.7cm]{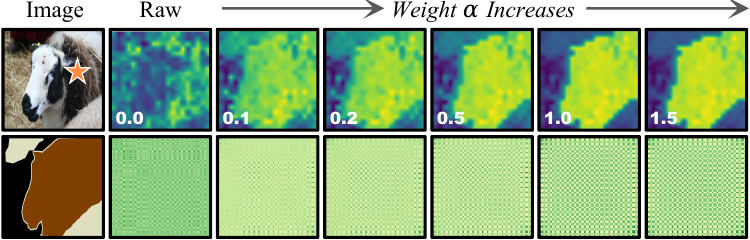}
   \caption{\hl{The effect of different $\alpha$ values on CLIP’s attention distribution, illustrating how SD's attention $\alpha$ enhances CLIP's attention.}}
   \label{fig.9}
   \vspace{-1em}
\end{figure}

In Table~{\ref{tab.6}} (e-g), we further investigate the effectiveness of our TSA module by varying the number of class centroids $E$ in Eq.~\ref{eq:9}, the number of learnable prompts $N$, and the cache weight $\beta$ in Eq.~\ref{eq:13}. To reduce the noise from generated images, we propose to cache the clustered image features as Keys. Table~{\ref{tab.6}} (e) shows that the performance drops when $E$ is either too small or too large. This could be because a small $E$ limits the knowledge stored in the cache model, while an excessively large $E$ introduces noisy features. DiCLIP achieves the best trade-off when $E=10$. Table~{\ref{tab.6}} (f) examines the impact of learnable prompts in our cache model, which are incorporated to enhance dynamic CAM generation. The results indicate that these learnable embeddings provide the most benefit when $N=92$ (where $312$ in brackets represents the total number of key-value pairs). In Table~{\ref{tab.6}} (g), we adjust the weight $\beta$ to balance the contribution of the cache CAM. When we decrease its contribution, the CAM performance drops from $78.2 \%$ to $75.9 \%$, further validating the efficacy of our cache model.
\begin{table*}[t!]
\centering
\caption{{Per-category performance analysis on VOC val set. The confusion ratio is used as the metric. We split the VOC val set into a single-class image set and a co-occurrence image set. 'avg.' means the average CR among all classes.}}
   \vspace{-1em}
    \tablestyle{2.15pt}{1}
    \scalebox{1.}
    {
    \footnotesize
    \begin{tabular}{l|cccccccccccccccccccccc}
    \toprule
    \multicolumn{23}{c}{Single-class image set}                                                                                                                                                                                                                                                                                                                                                                                                                                                                                                                                                                                                                                       \\ \midrule
     Method                                                              & bkg           & aero                                                         & bike          & bird                                                         & boat                                                         & bottle        & bus           & car           & cat                                                          & chair         & cow                                                          & table         & dog                                                          & horse                                                        & motor                                                        & person        & plant         & sheep                                                        & sofa                                                         & train                                                        & tv            & avg.                                                          \\ \midrule
    \multicolumn{1}{l|}{\textcolor{black}{AFA~\cite{12}~{\tiny CVPR'22}}}                                     & 0.03                        & 0.07                     & 2.42                     & 0.14                     & 0.20                      & 0.22                       & 0.07                     & 0.05                     & 0.19                     & 1.00                         & 0.06                     & 0.29                      & 0.10                      & 0.06                      & 0.15                      & 0.34                       & 0.44                      & 0.08                      & 0.45                     & 0.48                      & 0.30                      & 0.34                     \\
    \multicolumn{1}{l|}{\textcolor{black}{ToCo~\cite{22}~{\tiny CVPR'23}}}                                    & 0.02                        & 0.14                     & 0.96                     & 0.42                     & 1.16                     & \textbf{0.13}                       & 0.08                     & 0.09                     & 0.02                     & 0.81                      & 0.02                     & 0.50                      & 0.07                     & 0.02                      & 0.16                      & 0.05                       & 0.67                      & 0.06                      & \textbf{0.19}                     & 0.62                      & 0.28                     & 0.31                     \\
    \multicolumn{1}{l|}{\textcolor{black}{DuPL~\cite{DuPL}\tiny CVPR'24}}                                    & 0.02                        & 0.19                     & 1.07                     & 0.19                     & 0.53                     & 0.17                       & 0.08                     & 0.15                     & 0.03                     & 1.42                      & 0.01                     & 0.55                      & 0.06                     & 0.02                      & 0.10                      & 0.10                       & 0.56                      & 0.04                      & 0.43                     & 0.52                      & 0.94                     & 0.34                     \\
    \multicolumn{1}{l|}{\textcolor{black}{SeCo\cite{seco} \tiny CVPR'24}}                                    & {\color[HTML]{1A1C1E} 0.03} & 0.05                     & 1.35                     & 0.09                     & 0.32                     & 0.16                       & 0.05                     & \textbf{0.04}                     & 0.02                     & \textbf{0.75}                      & 0.01                     & 0.37                      & 0.09                     & 0.02                      & 0.09                      & 0.05                       & 0.30                       & 0.03                      & 0.26                     & 0.59                      & 0.35                     & 0.24                     \\
    \multicolumn{1}{l|}{\textcolor{black}{WeakCLIP~\cite{weakclip} \tiny IJCV'24}}                                & 0.02                        & 0.08                     & 0.94                     & 0.11                     & 0.20                     & 0.17                       & 0.06                     & 0.08                     & 0.04                     & 0.90                      & 0.02                     & 0.31                      & 0.08                     & 0.05                      & 0.13                      & 0.06                       & 2.27                      & 0.07                      & 0.65                     & 0.13                      & 0.59                     & 0.33                     \\
    \multicolumn{1}{l|}{\textcolor{black}{WeCLIP~\cite{18} \tiny CVPR'24}}                                  & 0.02                        & 0.08                     & 1.26                     & 0.07                     & 0.19                     & 0.15                       & 0.08                     & 0.09                     & 0.05                     & 0.98                      & 0.04                     & 0.40                      & \textbf{0.05}                     & 0.05                      & 0.10                      & 0.09                       & 0.45                      & 0.07                      & 0.62                     & 0.09                      & \textbf{0.18}                     & 0.24                     \\
    \multicolumn{1}{l|}{\textcolor{black}{MoRe~\cite{MoreZW}~\tiny AAAI'25}}                                    & 0.02                        & 0.10                     & 1.04                     & 0.06                     & 0.24                     & 0.14                       & 0.08                     & 0.07                     & \textbf{0.01}                     & 1.24                      & 0.01                     & 0.28                      & 0.06                     & 0.02                      & \textbf{0.08}                      & 0.15                       & \textbf{0.19}                      & \textbf{0.02}                      & 0.25                     & 0.50                      & 0.24                     & 0.22                     \\
    \rowcolor[HTML]{EFEFEF} 
    \multicolumn{1}{l|}{\cellcolor[HTML]{EFEFEF}\textbf{DiCLIP(Ours)}} & \textbf{0.02}               & \textbf{0.05}            & \textbf{0.65}            & \textbf{0.04}            & \textbf{0.13}            & 0.14                       & \textbf{0.04}            & 0.06                     & 0.04                     & 1.01                      & \textbf{0.01}            & \textbf{0.19}             & 0.07                     & \textbf{0.02}             & 0.10                      & \textbf{0.03}              & 0.31                      & 0.03                      & 0.40                     & \textbf{0.07}             & 0.40                     & \textbf{0.18}            \\ \midrule
    \multicolumn{23}{c}{Co-occurrence image set}                                                                                                                                                                                                                                                                                                                                                                                                                                                                                                                                                                                                                                                           \\ \midrule
     Method                                                              & bkg           & aero                                                         & bike          & bird                                                         & boat                                                         & bottle        & bus           & car           & cat                                                          & chair         & cow                                                          & table         & dog                                                          & horse                                                        & motor                                                        & person        & plant         & sheep                                                        & sofa                                                         & train                                                        & tv            & avg.                                                          \\ \midrule
    \multicolumn{1}{l|}{\textcolor{black}{AFA~\cite{12}~{\tiny CVPR'22}}}                                     & 0.08                        & 0.31                     & 1.83                     & 0.01                     & 0.28                     & \textbf{0.05}                       & 0.12                     & \textbf{0.09}                     & 0.28                     & 1.12                      & 0.20                     & 0.13                      & 0.20                     & 0.26                      & 0.38                      & 0.34                       & 0.27                      & 0.26                      & 0.64                     & 0.24                      & 0.89                     & 0.38                     \\
    \multicolumn{1}{l|}{\textcolor{black}{ToCo~\cite{22}~{\tiny CVPR'23}}}                                    & 0.07                        & 0.40                     & 0.68                     & 0.29                     & 0.95                     & 0.10                       & 0.14                     & 0.13                     & 0.04                     & 0.60                      & 0.10                     & 0.28                      & 0.11                     & 0.22                      & 0.25                      & 0.06                       & 0.45                      & 0.14                      & 1.17                     & 0.38                      & 0.55                     & 0.34                     \\
    \multicolumn{1}{l|}{\textcolor{black}{DuPL~\cite{DuPL} \tiny CVPR'24}}                                    & 0.05                        & 0.55                     & 0.92                     & 0.55                     & 0.55                     & 0.14                       & 0.12                     & 0.27                     & 0.08                     & 0.51                      & 0.09                     & 0.17                      & 0.14                     & \textbf{0.10}                      & 0.22                      & 0.10                       & 0.50                      & 0.16                      & 0.67                     & 0.26                      & 1.52                     & 0.37                     \\
    \multicolumn{1}{l|}{\textcolor{black}{SeCo\cite{seco} \tiny CVPR'24}}                                    & {\color[HTML]{1A1C1E} 0.07} & 0.16                     & 1.05                     & 0.29                     & 0.35                     & 0.21                       & 0.12                     & 0.12                     & 0.03                     & \textbf{0.41}                      & 0.04                     & 0.26                      & 0.08                     & 0.12                      & 0.24                      & 0.07                       & 0.27                      & 0.23                      & 0.40                      & 0.32                      & 0.69                     & 0.26                     \\
    \multicolumn{1}{l|}{\textcolor{black}{WeakCLIP~\cite{weakclip} \tiny IJCV'24}}                                & 0.05                        & 0.10                     & 0.93                     & 0.06                     & 0.29                     & 0.17                       & 0.10                     & 0.13                     & 0.03                     & 0.59                      & 0.05                     & 0.18                      & 0.13                     & 0.18                      & 0.22                      & 0.07                       & 2.14                      & 0.17                      & 0.40                     & 0.12                      & 1.01                     & 0.34                     \\
    \multicolumn{1}{l|}{\textcolor{black}{WeCLIP~\cite{18} \tiny CVPR'24}}                                  & 0.05                        & 0.15                     & 1.31                     & 0.04                     & 0.23                     & 0.15                       & 0.12                     & 0.17                     & 0.04                     & 0.75                      & 0.12                     & 0.15                      & 0.07                     & 0.22                      & 0.29                      & 0.11                       & 0.43                      & 0.18                      & 0.34                     & 0.13                      & \textbf{0.48}                     & 0.26                     \\
    \multicolumn{1}{l|}{\textcolor{black}{MoRe~\cite{MoreZW}~\tiny AAAI'25}}                                    & 0.05                        & 0.25                     & 1.02                     & 0.01                     & 0.21                     & 0.20                       & 0.10                     & 0.14                     & 0.07                     & 0.66                      & 0.11                     & 0.17                      & 0.09                     & 0.12                      & 0.23                      & 0.10                       & \textbf{0.22}                      & \textbf{0.09}                      & 0.54                     & 0.28                      & 0.73                     & 0.26                     \\
    \rowcolor[HTML]{EFEFEF} 
    \multicolumn{1}{l|}{\cellcolor[HTML]{EFEFEF}\textbf{DiCLIP(Ours)}} & \textbf{0.05}               & \textbf{0.09}            & \textbf{0.64}            & \textbf{0.01}            & \textbf{0.15}            & 0.17                       & \textbf{0.07}            & 0.10                     & \textbf{0.02}            & 0.79                      & \textbf{0.02}            & \textbf{0.12}             & \textbf{0.04}            & 0.16                      & \textbf{0.21}             & \textbf{0.05}              & 0.33                      & 0.10                      & \textbf{0.32}            & \textbf{0.09}             & 0.55                     & \textbf{0.19}           \\ \bottomrule
    \end{tabular}
    }
   \label{tab.8}
   \vspace{-0.5em}
\end{table*}

\begin{figure}[t]
  \centering
  \includegraphics[width=\linewidth]{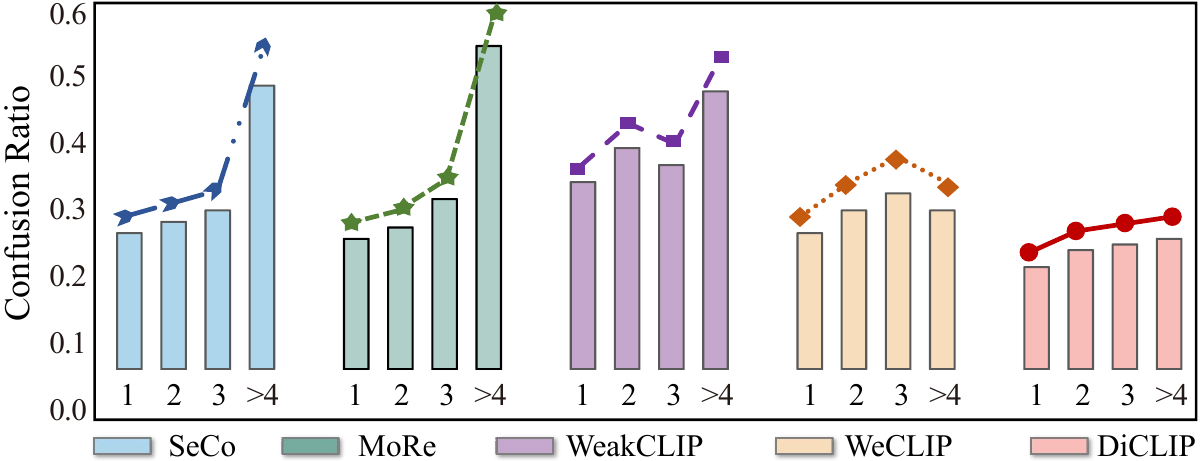}
  \vspace{-1.8em}
   \caption{\hl{Confusion ratio trends under increasing class co-occurrence on VOC val set. Our method achieves better performance in co-occurrence scenario.}}
   \label{fig.10}
   \vspace{-1.5em}
\end{figure}
Table~{\ref{tab.6}} (h) analyzes the influence of loss weight factor $\gamma$. It is observed that the segmentation performance drops when $\gamma$ is set to higher values. One reason is that the high weight factor of $\mathcal{L}_{ada}$ potentially undermines the role of segmentation loss. When set to a lower value, the performance keeps consistent as $\gamma$ varies. The above ablative studies convincingly demonstrate that DiCLIP is a robust WSSS to achieve significant performance under a wide range of settings.}

\begin{table}[!t]
    \centering
    \vspace{-0.5 em}
    \caption{Comparisons with fully-supervised methods on VOC val set. $\mathcal{C}$:Convolution-based. $\mathcal{T}$: Transformer-based. *: CLIP-based.}
    \vspace{-1em}
    \tablestyle{6.pt}{1.15}
    \scalebox{1.}
    {
    \footnotesize
    \begin{tabular}{l|c|c|cc}
    \toprule
    Methods        & Decoder                                                                                           & Backbone                                                            & Val           & Ratio           \\ \midrule
    \multicolumn{5}{l}{\textit{\textbf{Fully-supervised methods.}}}   \\ 

    {DeepLabV2~\cite{DeepLabV2} \tiny TPAMI'2017}      & $\mathcal{C}$                                                                     & \multicolumn{1}{c|}{RN101}                                      & 77.7          & -               \\
    {Segformer~\cite{segformer} \tiny NeurIPS'2021}      & $\mathcal{T}$                                                                    & \multicolumn{1}{c|}{MiT-B}                                   & 78.7          & -               \\
    {DeepLabV2~\cite{DeepLabV2} \tiny TPAMI'2017}      & $\mathcal{C}$                                                                    & \multicolumn{1}{c|}{ViT-B}                                   & 82.3          & -               \\
    {$\dagger$WeCLIP-Full~\cite{18} \tiny CVPR'2024}     & $\mathcal{C}$                                                                     & \multicolumn{1}{c|}{ViT-B*}                                  & 81.4          & -               \\
    {WeCLIP-Full~\cite{18} \tiny CVPR'2024}     & $\mathcal{T}$                                                                     & \multicolumn{1}{c|}{ViT-B*}                                  & 81.6          & -               \\ \midrule
    \multicolumn{5}{l}{\textit{\textbf{Weakly-supervised methods.}}}   \\ 
    {CLIMS~\cite{15} \tiny CVPR'2022}          & $\mathcal{C}$                                            & \multicolumn{1}{c|}{RN101}                                      & 70.4          & 90.6\%          \\
    {CLIP-ES~\cite{16} \tiny CVPR'2023}       & $\mathcal{C}$                                          & \multicolumn{1}{c|}{RN101}                                      & 72.2          & 92.9\%          \\
    {CPAL~\cite{37} \tiny CVPR'2024}          & $\mathcal{C}$                                          & \multicolumn{1}{c|}{RN101}                                      & 74.5          & 95.9\%          \\
    {ToCo~\cite{22} \tiny CVPR'2024}         &$\mathcal{C}$            & \multicolumn{1}{c|}{ViT-B}           & 71.1          & 86.4\%          \\
    {DuPL~\cite{DuPL} \tiny CVPR'2024}        & $\mathcal{C}$          & \multicolumn{1}{c|}{ViT-B}                                   & 73.3          & 89.1\%          \\
    {SeCo~\cite{seco} \tiny CVPR'2024}           & $\mathcal{C}$                                                        & \multicolumn{1}{c|}{ViT-B}                                   & 74.0          & 89.9\%          \\
    {WeakCLIP~\cite{18} \tiny IJCV'2024}         & $\mathcal{T}$         & \multicolumn{1}{c|}{MiT-B*}                                  & 75.1          & 95.4\%          \\
    {DIAL~\cite{DIAL} \tiny ECCV'2024}            & $\mathcal{C}$                                            & \multicolumn{1}{c|}{ViT-B*}                                   & 74.5          & 90.5\%          \\
    {WeCLIP~\cite{18} \tiny CVPR'2024}         & $\mathcal{T}$         & \multicolumn{1}{c|}{ViT-B*}                                  & 76.4          & 93.6\%          \\
    {\textcolor{black}{PCRE~\cite{PCRE} \tiny CVPR'2025}}         & $\mathcal{C}$         & \multicolumn{1}{c|}{ViT-B}                                  & \textcolor{black}{75.5}          & \textcolor{black}{91.7\%}          \\
    {\textcolor{black}{MoRe~\cite{MoreZW} \tiny AAAI'2025}}         & $\mathcal{C}$         & \multicolumn{1}{c|}{ViT-B}                                  & \textcolor{black}{76.4}          & \textcolor{black}{92.8\%}          \\
    \rowcolor[HTML]{EFEFEF} 
    {\cellcolor[HTML]{EFEFEF}\textbf{DiCLIP (Conv. Head)}} & \textbf{$\mathcal{C}$}  & \multicolumn{1}{c|}{\cellcolor[HTML]{EFEFEF}\textbf{ViT-B*}} & \textbf{78.7} & \textbf{96.7\%} \\ 
    
    \rowcolor[HTML]{EFEFEF} 
    {\cellcolor[HTML]{EFEFEF}\textbf{DiCLIP (Trans. Head)}} & \textbf{$\mathcal{T}$}  & \multicolumn{1}{c|}{\cellcolor[HTML]{EFEFEF}\textbf{ViT-B*}} & \textbf{78.8} & \textbf{96.6\%} \\  \bottomrule
    \end{tabular}
    }
   \label{tab.9}
   \vspace{-1.0em}
\end{table}
\subsection{Further Analysis}
{\textbf{{Per-category Performance Analysis.}}} {Table~{\ref{tab.8}} analyzes per-category performance trends in complex co-occurrence cases on VOC val set. We adopt Confusion Ratio (CR, computed as FP/TP; lower is better) metric~{\cite{seco}}. VOC val set is split into a single-class image set and a co-occurrence image set (images containing two or more classes). It reports that DiCLIP significantly outperforms recent SOTAs{~\cite{18}}, {\cite{weakclip}}, {\cite{seco}}, {\cite{MoreZW}} in the single-class set. Specifically, DiCLIP achieves $0.18$ CR across all categories, whereas the CLIP-based SOTA WeCLIP achieves $0.24$, indicating a $6\%$ improvement.

Notably, DiCLIP exhibits stronger superiority in challenging co-occurrence scenarios, where all competing methods experience a significant drop in performance on the co-occurrence set. In contrast, DiCLIP maintains a stable $0.19$ CR on the co-occurrence set. Furthermore, across the $21$ VOC classes, DiCLIP achieves superior performance in $14$ classes compared to recent competitors under these challenging settings.} {To provide a more intuitive view of failure trends, Fig.~{\ref{fig.10}} visualizes CR as a function of the number of co-occurring classes in each image set. As the number of co-occurring classes increases, most competing methods show a substantial rise in CR, whereas DiCLIP consistently maintains stable performance. Both the quantitative and qualitative results highlight DiCLIP's superiority in handling complex co-occurrence scenarios.}
\begin{figure*}
  \centering
  \includegraphics[width=18.1cm]{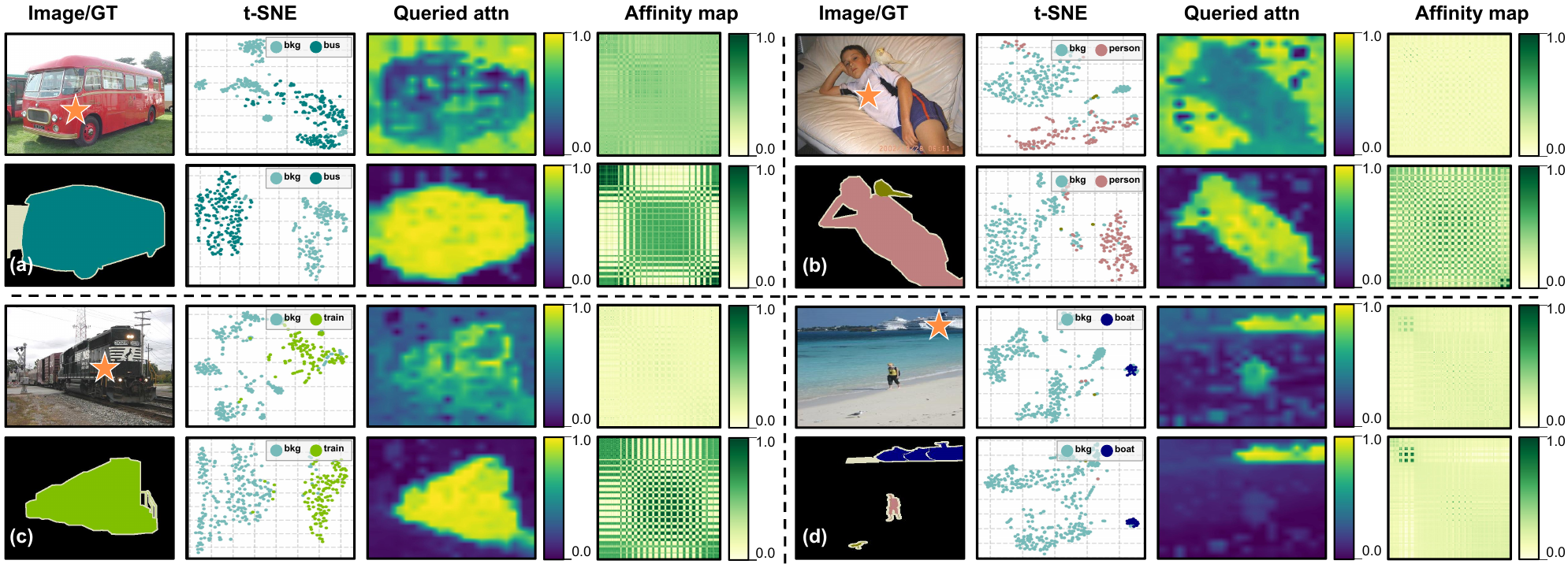}
   \caption{\hl{Visualization of feature representations. Three representative cases (a-c) and one small and off-center case (d) are illustrated. In each case, the upper row is the vanilla CLIP baseline, while the lower row shows ours. The t-SNE~{\cite{tsne}}, queried attention maps, and affinity maps are used for illustration.}}
   \label{fig.11}
  \vspace{-0.8em}
\end{figure*}

{\bf Comparison Fairness Analysis.} Considering backbone and decoder differences, Table~\ref{tab.9} evaluates WSSS methods against their fully-supervised counterparts to ensure fair comparisons. The SOTA WeCLIP uses a CLIP-based backbone and a Transformer-based decoder, which achieves $93.6 \%$ of its fully-supervised counterpart ($81.6\%$ mIoU). With the same settings, the proposed DiCLIP achieves $78.8\%$ mIoU, reaching $96.6\%$ of its fully-supervised counterpart, which significantly outperforms it by $3.0 \%$. To further explore the influence of decoder heads, we use the same Convolution-based decoder as~\cite{DIAL, DuPL}. The CLIP-based DIAL reaches $90.5 \%$ while our method achieves $96.7 \%$ of its counterparts. These comparison results fairly demonstrate the effectiveness of DiCLP in enhancing CLIP's dense potential with the diffusion model.

{\bf \textcolor{black}{Feature Representation Analysis.}} Fig.~\ref{fig.11} analyzes feature diversity when enhancing CLIP’s spatial awareness using the VCE module. The t-SNE~\cite{tsne} and two similarity-score-based techniques: queried attention maps and affinity maps, are utilized for illustration. For the t-SNE in each case, CLIP (upper row) confuses foreground and background features, whereas ours (lower row) generates a more distinctive space, clearly separating different semantics. Additionally, given a query patch token (marked with a yellow star), \hl{our queried attention maps exhibit high similarity with related semantics, even when the target objects are small and off-center (Fig.~{\ref{fig.11}} (d))}. It validates the efficacy of our module in enriching CLIP's visual knowledge with diverse priors from SD.

To further illustrate the impact of our enhancements on attention maps, we utilize affinity maps to measure pairwise token correlations, providing a clear assessment of the improved attention consistency. The visualization reveals that CLIP produces overly smoothed affinities, indicating limited spatial details. In contrast, DiCLIP generates more distinctive affinities. One reason is that we leverage diverse attention maps from SD (shown in Fig.~\ref{fig.3}) to diversify CLIP's suboptimal attention. \hl{These high-dimensional feature space comparisons provide further evidence for our first assumption (\emph{Spatial Consistency}) and validate the efficacy of the VCE module.}

\begin{table}[!t]
    \centering
    \vspace{-0.5em}
    \caption{Training efficiency comparisons on VOC train (CAM) and val set (Seg). All experiments are conducted on RTX 3090.}
    \vspace{-1em}
    \tablestyle{2.9pt}{1.27}
    \scalebox{1.}
    {
    \footnotesize
    \begin{tabular}{l|ccccc}
    \toprule
    Method                                  & Type                   & Required Time                               & GPU                                    & CAM                                   & Seg                                \\ \midrule
    {CLIMS~\cite{15} \tiny CVPR'2022}                                   & $\mathcal{M}$          & 1068 mins                          & 18.0 G                                   & 56.6                                  & 70.4                               \\
    {CLIP-ES~\cite{16} \tiny CVPR'2023}                                 & $\mathcal{M}$          & 420 mins                           & 12.0 G                                   & 70.8                                  & 72.2                               \\
    {MCTformer+~\cite{33} \tiny TPAMI'2024}                              & $\mathcal{M}$          & 1496 mins                          & 18.0 G                                   & 68.8                                  & 74.0                               \\
    {WeakCLIP~\cite{weakclip} \tiny IJCV'2024}                                  & $\mathcal{M}$          & 1160 mins                           & 9.7 G                                  & 73.7                                  & 75.1                               \\
   {AFA~\cite{12} \tiny CVPR'2022}                                     & $\mathcal{S}$          & 554 mins                           & 19.0 G                                 & 65.0                                  & 66.0                               \\
    {ToCo~\cite{22} \tiny CVPR'2023}                                    & $\mathcal{S}$          & 506 mins                           & 17.9 G                                 & 71.6                                  & 71.1                               \\
    {DuPL~\cite{DuPL} \tiny CVPR'2024}                                    & $\mathcal{S}$          & 508 mins                           & 14.9 G                                 & 75.0                                  & 73.3                               \\
    {SeCo~\cite{seco} \tiny CVPR'2024}                                    & $\mathcal{S}$          & 407 mins                           & 17.6 G                                 & 74.8                                  & 74.0                               \\
    {WeCLIP~\cite{18} \tiny CVPR'2024}                                  & $\mathcal{S}$          & 270 mins                           & 6.2 G                                  & 75.4                                  & 76.4                               \\
    \textcolor{black}{MoRe~\cite{MoreZW} \tiny AAAI'2025}                                  & $\mathcal{S}$          & \textcolor{black}{372 mins}                           & \textcolor{black}{12.1 G}                                  & \textcolor{black}{77.0}                                  & \textcolor{black}{76.4}   \\
    \rowcolor[HTML]{EFEFEF} 
    {\cellcolor[HTML]{EFEFEF}\textbf{{DiCLIP (Offline Phase)}}} & -                     & \textbf{35 mins}       & \textbf{4.8 G} & \textbf{74.0} & \textbf{-}    \\
    \rowcolor[HTML]{EFEFEF} 
    {\cellcolor[HTML]{EFEFEF}\textbf{{DiCLIP (Online Phase)}}}                          & \textbf{$\mathcal{S}$} & \textbf{80 mins}                   & {7.5 G}                         & \textbf{78.2}    & \textbf{78.8}                      \\ \bottomrule
    \end{tabular}
    }
   \label{tab.10}
   \vspace{-1.8 em}
\end{table}

{\bf Training Feasibility Analysis.} DiLCIP achieves significant performance based on a key-value cache model. One concern is whether this precondition hinders its feasibility. We investigate it from training efficiency and convergence speed. {Table{~\ref{tab.10}} analyzes the training efficiency of our approach compared to recent SOTAs. ‘DiCLIP (Offline Phase)’ refers to the cache construction process, whereas ‘DiCLIP (Online Phase)’ only reflects the training period (excluding cache building). It reports that building a cache takes $35$ minutes with $4.8$ GB GPU memory.} With these acceptable training costs, DiCLIP is competent to generate high-quality CAM with $74.0 \%$ mIoU in a training-free manner. {In addition, DiCLIP (Online Phase) can further train WSSS pipeline in a single-stage way.} Our method only requires $80$ minutes with $7.5$ GB GPU memory, and generates $78.2 \%$ and $78.8 \%$ mIoU for CAM seeds and segmentation maps. The whole process takes $115$ minutes, requiring only $42.6 \%$ and $9.9 \%$ training time of single-stage SOTA WeCLIP and multi-stage SOTA WeakCLIP. 

In addition, Fig.~\ref{fig.12} illustrates the convergence speed by showing the quality of CAM seeds at different training iterations. When DiCLIP is initiated with the cache prior, it quickly reaches a higher performance plateau with only $4k$ iterations and remains stable afterward, whereas the baseline methods, WeCLIP and WeakCLIP, require $8k$ and $16k$ iterations to achieve convergence. This is because our cache model stores rich category knowledge, enabling DiCLIP to quickly learn reliable category representations for WSSS. These results in training efficiency and convergence speed underscore the practical feasibility of our method in real-world applications.
\begin{figure}[!t]
  \centering
  \includegraphics[width=\linewidth]{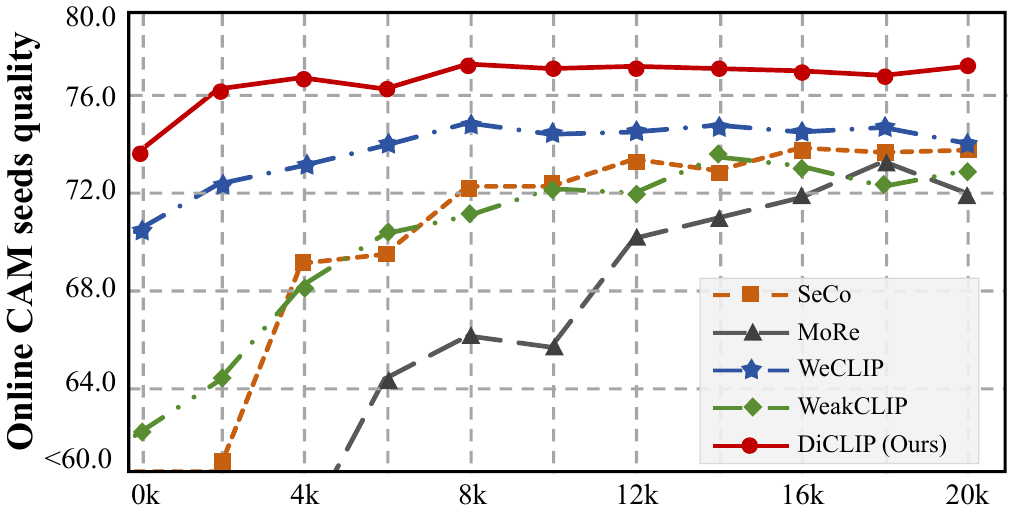}
  \vspace{-1.8em}
   \caption{\hl{Comparison of convergence speed during training on the VOC validation set, showing that our method converges faster than baselines.}}
   \label{fig.12}
   \vspace{-1.em}
\end{figure}

\section{{Further Discussion}}
\subsection{{Comprehensive Discussion of ACR Module.}} 

{\textbf{A.1) Rationale for Applying ACR on SD.} ACR module refines attention by clustering spatially diverse and semantically meaningful patterns. This approach is well-suited for SD attention maps, as they display clear and localized focus regions, as illustrated in Fig.{~\ref{fig.13}} (b), making them amenable to clustering-based refinement. To further validate this, we directly apply ACR to CLIP's attention. As shown in Fig.{~\ref{fig.13}} (a), CLIP’s attention maps are inherently smooth and low-contrast, leading to unstable clustering and amplification of incorrect semantics. The quantitative results in Table{~\ref{tab.11}} (a) confirm this observation, where directly applying ACR to CLIP yields patch-text CAM $M_t$ with only $22.1\%$ mIoU on the VOC train set, indicating poor performance. In contrast, applying ACR to SD achieves significant improvement, supporting the rationale of ACR for SD-based attention refinement.}

{\textbf{A.2) Necessity of Clustering-based Refinement.} The ACR module adopts a clustering-based strategy to diversify SD's attention. A natural concern is whether a threshold-based refinement could serve as a simpler alternative. To investigate it, we compare the two strategies. As shown in Table{~\ref{tab.11}} (b), ACR achieves more favorable $M_t$ performance on VOC train set. The advantages of our approach lie in two aspects: its ability to control granularity and to preserve subtle but important information. In our ACR, the number of centroids $B$ explicitly controls the partitioning granularity, enabling fine-grained and iterative refinement for coherent semantic regions. In contrast, thresholding only removes values below a fixed cutoff (the mean SD attention), and lacks a mechanism to adjust regional detail. Moreover, clustering assigns all attention values to centroids, thereby retaining weak but semantically meaningful responses. Thresholding, however, discards low-intensity values and may lose subtle cues.}

\begin{figure}[t]
  \centering
  \includegraphics[width=8.7cm]{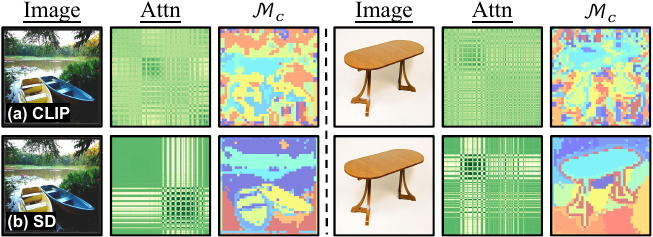}
   \caption{\hl{Comparison between ACR on CLIP's attention and on SD's attention, showing that SD’s localized attention enables more effective clustering.}}
   \label{fig.13}
  \vspace{-.5em}
\end{figure}
\begin{table}[t]
  \centering
  \caption{\hl{Discussion of ACR module with $M_t$ on VOC train set, validating the superiority of our module.}}
  \vspace{-1.5em}
  \subfloat[{ACR ON CLIP VS. SD ATTENTION.}]
  {
    \centering
    \begin{minipage}[b]{0.45\linewidth}{
    \begin{center}
    \tablestyle{2pt}{1.05}
    \scalebox{1.0}{
    \begin{tabular}{l|ccc}
    \toprule
    Conditions  & Precision     & Recall        & mIoU          \\ \midrule
    On CLIP        & 30.2          & 47.3          & 22.1          \\
    \rowcolor[HTML]{EFEFEF} 
    \textbf{On SD} & \textbf{83.9} & \textbf{83.0} & \textbf{72.0} \\ \bottomrule
    \end{tabular}
    }
    \end{center}
    }
    \end{minipage}
   }
\subfloat[{ACR VS. THRESH-BASED REFINE.}]
  {
    \centering
    \begin{minipage}[b]{0.55\linewidth}{
    \begin{center}
    \tablestyle{2pt}{1.05}
    \scalebox{1.0}{
    \begin{tabular}{l|ccc}
    \toprule
    Conditions   & Precision     & Recall        & mIoU          \\ \midrule
    Thresh.       & 83.7          & 82.5          & 71.5          \\
    \rowcolor[HTML]{EFEFEF} 
    \textbf{ACR} & \textbf{83.9} & \textbf{83.0} & \textbf{72.0} \\ \bottomrule
    \end{tabular}
    }
    \end{center}
    }
    \end{minipage}
  }
  \label{tab.11}%
   \vspace{-1.em}
\end{table}%
\begin{table}[t!]
\centering
\caption{{Comparisons of fusion strategies with $M_t$ on VOC train set.}}
   \vspace{-1em}
    \tablestyle{7.6pt}{1}
    \scalebox{1.}
    {
    \footnotesize
    \begin{tabular}{l|c|c|ccc}
    \toprule
    Conditions    & Single & Fusion & Precision     & Recall        & mIoU          \\ \midrule
    Only CLIP        & \pmb{$\checkmark$}      &         & {21.0} & 26.8          & 12.9          \\
    Only SD        & \pmb{$\checkmark$}      &         & {84.0} & 82.1          & 71.4          \\ \midrule
    Gating        &       & \pmb{$\checkmark$}        & {84.1} & 81.0          & 70.5          \\
    Scaling       &       & \pmb{$\checkmark$}        & 44.1          & 55.2          & 33.1          \\
    Normalization &       & \pmb{$\checkmark$}        & 15.9          & 20.7          & 10.0          \\
    \rowcolor[HTML]{EFEFEF} 
    \textbf{Ours} &        & \pmb{$\checkmark$}      & 83.9          & \textbf{83.0} & \textbf{72.0} \\ \bottomrule
    \end{tabular}
    }
   \label{tab.12}
   \vspace{-1.5em}
\end{table}
{\textbf{A.3) Different Fusion Strategies.} Analogous to the principle of LoRA~{\cite{lora}}, we adopt a direct additive fusion strategy, where the SD attention map serves as a meaningful bias to enrich the relatively smooth CLIP attention. By adding SD attention to CLIP’s, we encourage more diverse attention responses while preserving CLIP’s inherent image–text alignment, thereby benefiting CAM generation. To further assess this, we directly replace CLIP’s attention with SD’s. As reported in Table{~\ref{tab.12}}, this naive replacement (Only SD), while enhancing CLIP’s dense capability, simultaneously weakens its cross-modal consistency. Consequently, it leads to inferior $M_t$ performance on VOC train set, which underscores the superiority of our proposed additive fusion strategy.

We also conduct a comprehensive comparison between our additive fusion and alternative fusion strategies—normalization, scaling, and gating—as reported in Table{~\ref{tab.12}}. Normalization applies an additional softmax after adding SD attention, but this over-smooths the SD distribution and weakens its corrective effect. Scaling multiplies CLIP’s attention by the SD map, which fails to enhance spatial diversity and severely distorts CLIP’s feature distribution. Gating sets a threshold (the mean SD attention), replacing only the most confident SD activations. While it improves the performance, it removes parts of CLIP’s original attention and discards subtle SD cues, undermining both alignment and diversity. \hl{In contrast, our additive fusion, though simple, preserves CLIP’s inherent alignment and SD’s complementary bias, resulting in better performance.}
}
\begin{figure}[t]
  \centering
  \includegraphics[width=\linewidth]{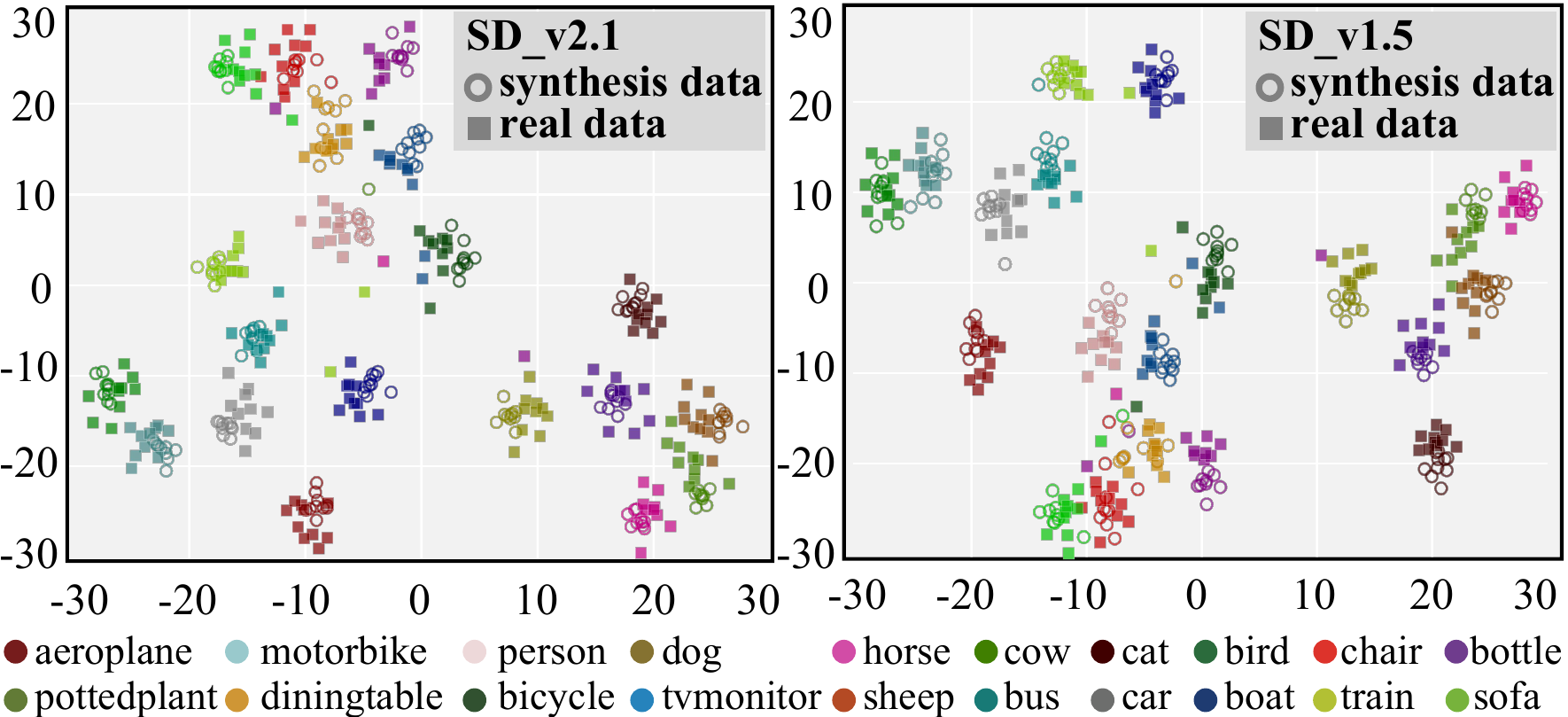}
   \caption{\hl{t-SNE visualization of cache feature distributions for synthetic data from SD v$2.1$ and v$1.5$ versus real VOC images, showing that synthetic distributions are close to the real data.}}
   \label{fig.14}
   \vspace{-1em}
\end{figure}

\subsection{{Comprehensive Discussion of Synthetic Data Bias.}} 

{DiCLIP generates the key-value cache with SD-generated images, which may inherit biases from the pre-trained diffusion model. To investigate it, we present three key findings: 

\textbf{B.1) Distribution Similarity.} We use t-SNE to compare cache distributions between synthetic and real data. The real cache is constructed using 50 images per category from the VOC train set, supplemented with COCO when single-class samples are insufficient. As shown in Fig.~\ref{fig.14}, synthetic and real data exhibit similar feature distributions, benefiting from SD’s strong generative ability and CLIP’s generalization across diverse images. Notably, this similarity persists across different SD models, indicating that the synthetic data consistently resembles real data across model variants.

\hl{\textbf{B.2) Semantic Purity as an Implicit Noise Filter.}  We generate single-category images with clear foreground objects. This design preserves the semantic purity of the synthetic cache, which could serve as a critical noise filter during retrieval. When applied to real images, the semantically pure prototypes in our cache model reinforce category-related cues and suppress spurious background correlations, enabling DiCLIP to better disentangle target objects from cluttered scenes. This analysis is consistent with the superior performance observed on real-world co-occurrence scenarios in Table{~\ref{tab.8}}.}

\begin{table}[t!]
\centering
\caption{{Comparisons between caches built from the VOC dataset ( Cache\_V) and from SD across various versions (Cache\_S).}}
   \vspace{-1em}
    \tablestyle{6.05pt}{1}
    \scalebox{1.}
    {
    \footnotesize
    \begin{tabular}{l|cc|cccc}
    \toprule
    Conditions       & Cache\_V & Cache\_S & $M_t$          & $M_{es}$         & $M_{ed}$         & Seg           \\ \midrule
    Real\_Data       & $\checkmark$          &           & 72.0          & \textbf{74.3} & 78.2          & \textbf{77.2} \\
    SD\_v1.4          &            & $\checkmark$         & 71.6          & 73.4          & 77.7          & 76.7          \\
    SD\_v1.5          &            & $\checkmark$         & 71.9          & 73.8          & 78.0          & 76.9          \\
    \rowcolor[HTML]{EFEFEF} 
    \textbf{SD\_v2.1} &            &$\checkmark$         & \textbf{72.0} & 74.0          & \textbf{78.2} & 77.1          \\ \bottomrule
    \end{tabular}
    }
   \label{tab.13}
   \vspace{-1em}
\end{table}
\begin{table}[t!]
\centering
\caption{{Impact of the number of generated images on $M_{es}$ (VOC train).}}
   \vspace{-1em}
    \tablestyle{9.6pt}{1}
    \scalebox{1.}
    {
    \footnotesize
    \begin{tabular}{c|cccccc}
    \toprule
    Nums & $M_t$ & 10   & 20   & 40   & {50}   & 100  \\ \midrule
    mIoU & 72.0     & 72.9 & 73.5 & 73.9 & {74.0} & 74.1 \\ \bottomrule
    \end{tabular}
    }
   \label{tab.14}
   \vspace{-1em}
\end{table}
\begin{table}[t!]
\centering
\caption{{Cross-dataset evaluation of segmentation on VOC and COCO.}}
   \vspace{-1em}
    \tablestyle{6pt}{1}
    \scalebox{1.}
    {
    \footnotesize
    \begin{tabular}{l|cccccc}
    \toprule
    Percentage & VCE  & 0C   & 25\%C & 50\%C & 75\%C & 100\%C \\ \midrule
    VOC          & 72.2 & 74.6 & 75.8  & 76.5  & 76.8  & 77.1   \\
    COCO         & 41.3 & 42.1 & 43.9  & 44.7  & 46.8  & 47.6   \\ \bottomrule
    \end{tabular}
    }
   \label{tab.15}
   \vspace{-1em}
\end{table}

\textbf{B.3) Bias Mitigation via Dynamic Dense Retrieval.} Our DiCLIP highlights a dynamic dense retrieval process, which adapts our model to the downstream domains and alleviates the domain bias. To investigate this, we compare caches built from VOC dataset and from SD across various versions. As shown in Table~\ref{tab.13}, although caches built from VOC yield higher cache CAM $M_{es}$, initializing the adapter with SD-based caches effectively compensates for the performance drop, validating DiCLIP’s ability to alleviate domain bias.

\textbf{B.4) Practical Advantages.} It should be noted that collecting single-category real images for a cache model is highly challenging. For instance, the VOC training set with $10,582$ images contains only $17$ single-class samples for the Dining-Table category. The difficulty becomes even more pronounced in complex scenarios where the single-class images are not available, highlighting the practicality and superiority of using SD-generated images for cache construction.
}

\subsection{\hl{Comprehensive Discussion of Robustness\&Generalization.}}

\hl{\textbf{C.1) Robustness to Different SD Versions.}} The proposed DiCLIP adopts SD with a version $2.1$ to enhance CLIP's dense capability. In Table{~\ref{tab.13}}, we further validate its performance across different SD versions, including SD\_v1.4, SD\_v1.5, and SD\_v2.1. The results show that DiCLIP consistently maintains robust performance across different SD versions, benefited from SD’s strong generative ability and CLIP’s generalization towards different types of images.

\hl{\textbf{C.2) Robustness to the Number of Generated Images.}} In Table{~\ref{tab.14}}, we further evaluate our robustness to different numbers of synthetic samples. As the number increases, the performance gradually improves. The results show that even with only $10$ generated images per category, DiCLIP improves patch-text CAMs by $0.9\%$ mIoU on the VOC train set. \hl{This confirms that our method is robust to data sparsity and further supports our second assumption (\emph{Generative Validity}).}
 
\hl{\textbf{C.3) Cross-dataset Generalization.}} In Table{~\ref{tab.15}}, we validate the generalization of DiCLIP on both VOC and COCO datasets. ‘0C’ denotes key–value pairs are not built and the adapter is randomly initialized, relying solely on $M_{t}$ for supervision. DiCLIP achieves $74.6\%$ mIoU on VOC and shows consistent improvements on COCO, demonstrating the adapter’s strong adaptability. ‘25\%C’ indicates that 25\% of categories are used to build a cache. Under this setting, the performance on VOC increases to 75.8\% mIoU, reaching comparable SOTA levels. \hl{Across both datasets, the improvements highlight that the key–value cache from seen classes provides transferable prior knowledge and generalizes well to new categories.}

The comprehensive analyses on the impact of SD versions, the number of generated images, and cross-dataset evaluations further validate the remarkable performance of DiCLIP, underscoring its strong potential in more challenging scenarios.

\section{Conclusion}
\label{sec:conclusions}
In this paper, we propose DiCLIP, a novel WSSS framework that leverages Stable Diffusion to enhance CLIP’s capability for CAM generation. We show that the spatial consistency and generative power of diffusion models effectively boost CLIP’s dense knowledge across both visual and text modalities. To this end, we introduce the Visual Correlation Enhancement (VCE) module to diversify CLIP’s visual features using SD attention, and the Text Semantic Augmentation (TSA) module to enrich text embeddings via a visual key–value cache, establishing a patch-wise knowledge retrieval paradigm for CAM generation. With these enhancements, DiCLIP surpasses existing SOTA methods while significantly reducing training costs. Extensive experiments validate its effectiveness and potential for efficient WSSS, providing a foundation for future exploration of CLIP and diffusion models in the WSSS community.

\hl{One limitation of DiCLIP is that the key–value cache is generated solely from SD, which may introduce biases from synthetic data. While our dynamic knowledge retrieval mitigates this issue, the method may underperform for categories poorly represented by the diffusion model or in highly complex scenes, as reflected in the failure cases (Fig.~{\ref{fig.5}}). A more direct solution is to progressively incorporate queried images and dynamically maintain an online cache during training, integrating prior knowledge with in-domain knowledge from the real data. In future work, we aim to develop such an online cache that adaptively stores category information from test samples, thereby further enhancing both performance and efficiency of the WSSS framework.}


\bibliographystyle{IEEEtran}
\bibliography{IEEEabrv,references}



\end{document}